\documentclass{article}

\PassOptionsToPackage{numbers, compress}{natbib}

\usepackage[final]{neurips_2024}

\usepackage[utf8]{inputenc} 
\usepackage[T1]{fontenc}    
\usepackage{hyperref}       
\usepackage{url}            
\usepackage{booktabs}       
\usepackage{amsfonts}       
\usepackage{nicefrac}       
\usepackage{microtype}      
\usepackage{xcolor}         

\usepackage{amsmath,amsfonts,bm}

\def\eqref#1{equation~\ref{#1}}

\def\1{\bm{1}}

\def\ve{{\bm{e}}}

\def\vs{{\bm{s}}}
\def\vt{{\bm{t}}}

\def\vw{{\bm{w}}}
\def\vx{{\bm{x}}}

\def\vz{{\bm{z}}}

\DeclareMathAlphabet{\mathsfit}{\encodingdefault}{\sfdefault}{m}{sl}
\SetMathAlphabet{\mathsfit}{bold}{\encodingdefault}{\sfdefault}{bx}{n}

\def\gI{{\mathcal{I}}}

\def\gM{{\mathcal{M}}}

\def\gT{{\mathcal{T}}}

\usepackage{wrapfig}
\usepackage{multirow}
\usepackage{makecell}
\usepackage{amsmath}
\usepackage{capt-of}
\usepackage{tabularx}
\usepackage{epsfig}
\usepackage{amssymb}
\usepackage{amsfonts}
\usepackage{booktabs}
\usepackage{scalerel}
\usepackage[inline]{enumitem}
\usepackage{listings}
\usepackage{varwidth}
\usepackage[export]{adjustbox}
\usepackage{tikz}
\usetikzlibrary{tikzmark}

\usepackage{stmaryrd}
\usepackage{bbm}
\usepackage{wrapfig}
\usepackage{pifont}

\definecolor{deepblue}{rgb}{0,0,0.5}
\definecolor{officeblue}{RGB}{0,102,204}
\definecolor{deepred}{rgb}{0.6,0,0}
\definecolor{deepgreen}{rgb}{0,0.5,0}
\definecolor{mybrickred}{RGB}{182,50,28}

\definecolor{fillcolor}{RGB}{216,217,252}

\newcommand*\AlgCommentInLine[1]{{\color{deepblue}{$\triangleright$ \textit{#1}}}}

\usepackage{pifont}

\usepackage{graphicx}
\usepackage{amsmath}

\usepackage{amssymb}

\usepackage{booktabs}
\usepackage{amsthm,amsmath,amssymb}
\usepackage{mathrsfs}

\usepackage{bbding}

\usepackage{amsmath}
\DeclareFontFamily{U}{mathc}{}
\DeclareFontShape{U}{mathc}{m}{it}%
{<->s*[1.03] mathc10}{}

\DeclareMathAlphabet{\mathscr}{U}{mathc}{m}{it}

\usepackage{array, tabularx, boldline}
\usepackage{graphicx}
\usepackage{cellspace}
\usepackage{color}

\usepackage{epsfig}
\usepackage{graphicx}
\usepackage{amsmath}
\usepackage{amssymb}
\usepackage{amsthm}
\usepackage{algpseudocode,algorithm,algorithmicx}
\usepackage{algpseudocode}
\usepackage[algo2e]{algorithm2e}  

\usepackage{graphicx}
\usepackage{amsmath}
\usepackage{amssymb}
\usepackage{booktabs}
\usepackage{float}
\usepackage{pifont}
\usepackage{enumitem}
\usepackage{multirow}

\definecolor{myblue}{rgb}{0.27, 0.80, 1.0}
\definecolor{mygreen}{rgb}{0.6, 1.0, 0.6}
\definecolor{myred}{rgb}{1.0, 0.2, 0.2}

\usepackage{makecell}
\usepackage{colortbl}
\usepackage{booktabs}
\usepackage{multirow}

\usepackage{bm}

\hypersetup{
    pagebackref,
    colorlinks=true,
    linkcolor=purple,
    filecolor=black,      
    citecolor=cyan, 
}

\newlength\secmargin
\newlength\subsecmargin
\newlength\paramargin
\newlength\figmargin
\newlength\eqmargin
\usepackage[capitalize]{cleveref}
\crefname{section}{Sec.}{Secs.}
\Crefname{section}{Section}{Sections}
\Crefname{table}{Table}{Tables}
\crefname{table}{Tab.}{Tabs.}
\crefname{algorithm}{Algo.}{Algos.}

\newcommand{\ie}{\textit{i}.\textit{e}.}
\newcommand{\eg}{\textit{e}.\textit{g}.}
\newcommand{\etc}{\textit{etc}.~}

\title{OneRef: Unified One-tower Expression Grounding and Segmentation with Mask Referring Modeling}

\author{%
  Linhui Xiao$^{1,2,3}$, Xiaoshan Yang$^{1,2,3}$, Fang Peng$^{1,2,3}$, Yaowei Wang$^{2,4}$, Changsheng Xu$^{1,2,3}$\thanks{Corresponding author.}  \\
  $^{1}$MAIS, Institute of Automation, Chinese Academy of Sciences \quad $^{2}$Pengcheng Laboratory\\
  $^{3}$School of Artificial Intelligence, University of Chinese Academy of Sciences \\
  $^{4}$Harbin Institute of Technology (Shenzhen) \\
  \texttt{\{xiaolinhui16, pengfang21\}@mails.ucas.ac.cn}, \\
  \texttt{\{xiaoshan.yang, csxu\}@nlpr.ia.ac.cn}, \texttt{wangyw@pcl.ac.cn}  \\
}

\begin{document}

\maketitle

\begin{abstract}
Constrained by the separate encoding of vision and language, existing grounding and referring segmentation works heavily rely on bulky Transformer-based fusion en-/decoders and a variety of early-stage interaction technologies. Simultaneously, the current mask visual language modeling (MVLM) fails to capture the nuanced referential relationship between image-text in referring tasks. In this paper, we propose \textit{\textbf{OneRef}}, a minimalist referring framework built on the modality-shared one-tower transformer that unifies the visual and linguistic feature spaces. To modeling the referential relationship, we introduce a novel MVLM paradigm called \textit{Mask Referring Modeling (\textbf{MRefM})}, which encompasses both referring-aware mask image modeling and referring-aware mask language modeling. Both modules not only reconstruct modality-related content but also cross-modal referring content. Within MRefM, we propose a referring-aware dynamic image masking strategy that is aware of the referred region rather than relying on fixed ratios or generic random masking schemes. By leveraging the unified visual language feature space and incorporating MRefM's ability to model the referential relations, our approach enables direct regression of the referring results without resorting to various complex techniques. Our method consistently surpasses existing approaches and achieves SoTA performance on both grounding and segmentation tasks, providing valuable insights for future research. Our code and models are available at \url{https://github.com/linhuixiao/OneRef}.
\end{abstract}

\section{Introduction}
\label{1.0-introduction}

Visual Grounding (VG) aims to ground a region referred by a expression query text in a specific image. The generalized VG / referring tasks include Referring Expression Comprehension (REC) \cite{qiao2020referring,mao2016generation,yu2016modeling,hu2016natural,deng2021transvg, xiao2023clip, xiao2019dynamic, li2019integrated}, Phrase Grounding (PG) \cite{akbari2019multi, rohrbach2016grounding}, and Referring Expression/Image Segmentation (RES/RIS) \cite{qiao2020referring, lavt, wang2022cris}. In REC/PG, the grounding region is represented by a rectangular boundary box, while in RES/RIS, it is represented by an irregular fine-grained segmented mask of the referred object. Unlike object detection \cite{liu2023cigar, liu2023foregroundness} or instance segmentation \cite{he2017mask}, which usually relies on a close-set of categories to detect or segment multiple regions that satisfy the object label, visual grounding is not limited to fixed categories. It requires understanding the semantics of the query text and then grounding or segmenting specific areas. Therefore, visual grounding is a task that strongly relies on the multimodal interaction and alignment of visual and linguistic features.

\begin{figure*}[t]
\vspace{-6pt}
 \centering
   \includegraphics[width=1.0\linewidth]{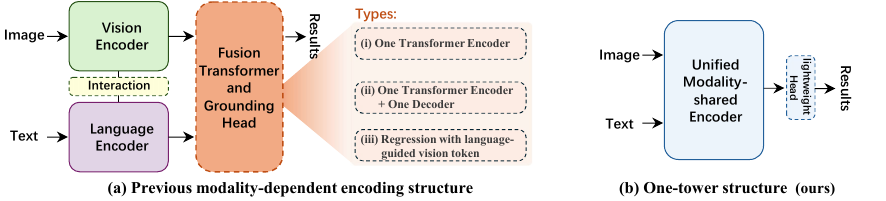}
   \vspace{-18pt}
   \caption{Comparison between our proposed approach and the mainstream REC/RES architectures.}
    \vspace{-17pt}
   \label{fig:fig1}
\end{figure*}

Since the introduction of BERT \cite{bert} and ViT \cite{vit, detr}, the state-of-the-art (SoTA) grounding works have widely adopted a pre-training and fine-tuning paradigm. As illustrated in \cref{fig:fig1}, existing studies employing pre-trained models, either utilizing uni-modal pre-trained models to separately transfer visual and language knowledge \cite{deng2021transvg, transvg++, qrnet, kamath2021mdetr, liu2023grounding} or utilizing multimodal pre-trained models \cite{xiao2023clip, shi2023dynamic, wang2022cris, kim2023risclip}, primarily fall into three typical architectures: \textbf{\textit{(i)}} two modality encoders combined with a cross-modal fusion encoder, exemplified by TransVG \etc \cite{chen2020uniter, deng2021transvg, qrnet, wang2022cris, xiao2023clip, kim2023risclip, xiao2024hivg}; \textbf{\textit{(ii)}} additionally incorporating a decoder, exemplified by MDETR \etc \cite{kamath2021mdetr, li2021referring, yang2022unitab, wang2022ofa, liu2023dqdetr, liu2023polyformer, shi2023dynamic, liu2023grounding}; \textbf{\textit{(iii)}} direct regression based on language-guided visual features, such as LAVT, TransVG++, \etc \cite{lavt, transvg++, vg-law, ye24uniqrnet}. However, incorporating modality-dependent encoders in these studies presents a challenge for seamlessly integrating the two modalities into a unified feature space. Consequently, these works not only require an additional cross-modal Transformer-based \cite{transformer} en-/decoder (\textit{(i)} and \textit{(ii)}), but also propose a variety of careful-designed interaction structures for modality-dependent encoders to facilitate early-stage fine-grained cross-modal alignment \cite{transvg++, qrnet, xiao2024hivg, vg-law, liu2023dqdetr, kim2023risclip, liu2023grounding,liu2023fdo}, such as adapter \cite{transvg++, kim2023risclip}, cross-modal bridge \cite{xiao2024hivg}, weight generation \cite{vg-law}, image-text cross-attention \cite{liu2023grounding, liu2023dqdetr}, \etc Therefore, these methods not only entail a large number of parameters but also involve intricate processes. Considering these critical limitations, we aim to explore simpler modality-shared grounding frameworks that can unify vision and language within a unified feature space, thereby obviating the necessity of the elaborate interaction modules, bulky fusion Transformer en-/decoders, as well as the special grounding tokens.

With the advancement of pre-training \cite{radford2021learning, peng2023sgva}, several studies have been conducted to explore unified modality-shared multimodal frameworks. YORO \cite{ho2023yoro} implemented a shared encoder based on ViLT \cite{kim2021vilt}. However, its modeling approach tends to overshadow the uni-modal knowledge and requires the encoder to incorporate additional query anchors, limiting its applicability for transfer with common pre-trained models. ONE-PEACE \cite{wang2023one} has designed seven expert branches based on Mix-of-Expert (MoE) \cite{bao2022vlmo, shazeer2016outrageously, fedus2022switch} to construct a three-modality foundation model to realize the integration of image, text, and audio modalities. However, their research employed extensive tri-modal data without exploring the potential utilization of MVLM for modeling the referring tasks. BEiT-3 \cite{beit3} is built upon multi-way Transformer \cite{bao2022vlmo, tang2022mmt}, which adopts three MoE heads (\ie, vision, language, vision-language) and a modality-shared structure that effectively unifies vision and language within a shared feature space. It demonstrates notable advantages across various classification-like cross-modal fields (\eg, Retrieval, VQA \etc). However, no prior research has explored the utilization of BEiT-3 for achieving transfer in referring tasks. Consequently, our objective is to explore more concise and efficient referring grounding and segmentation transfer within a unified feature space on the one-tower model of BEiT-3. However, BEiT-3 model is pre-trained utilizing a generic Mask Vision Language Modeling (MVLM) approach, and this masking paradigm lacks fine-grained cross-modal referring ability and cannot effectively model the intricate referential relationship between images and text. As a result, there exists a significant gap when applying BEiT-3 to the regression-like referring tasks. Therefore, exploring how to incorporate fine-grained cross-modal referring capability into the mask modeling paradigm becomes an important research issue that has not been addressed yet.

In this paper, we propose a novel paradigm called \textit{\textbf{Mask Referring Modeling} (\textbf{MRefM})}, as well as a unified and extremely concise grounding and referring segmentation framework named \textit{\textbf{OneRef}} that no longer requires the fusion or interaction Transformer structure and the special grounding tokens.

\textbf{Firstly}, we propose MRefM paradigm to enhance the referring capability of BEiT-3 in a flexible manner. MRefM consists of two components: Referring-aware Mask Image Modeling (\textbf{Referring MIM}) and Referring-aware Mask Language Modeling (\textbf{Referring MLM}). The conventional MVLM is typically trained alternately or randomly with uni-modal MIM and MLM. In contrast, Referring MIM and Referring MLM are required to reconstruct two distinct types of content: their own modality-related content and cross-modal referring information. Specifically, \textit{\textbf{(i) Referring MIM}} employs visual tokens after the dot product operation with the aggregated text token for reconstruction purposes. This not only entails reconstructing masked visual features itself but also necessitates reconstructing the visual target-relation score, which indicates the distance between the current token and the grounding region. The score encompasses four dimensions: horizontal and vertical distance to the grounding center, as well as width and height of the grounding region. In order to enhance the model's understanding capability for referred regions, we propose a referring-aware dynamic image masking strategy that replaces traditional ratio-fixed random masking so that referred regions are reconstructed with a relatively high mask ratio. \textit{\textbf{(ii) Referring MLM}} employs text tokens after the dot product operation with the aggregated visual token for reconstruction purposes. This not only involves reconstructing masked text itself but also requires reconstructing semantic target-relation scores that represent the correlation degrees between current text tokens and referred image regions.

\textbf{Secondly}, existing grounding and segmentation models commonly employ a \texttt{[Region]} token and multiple query anchors to regress results. However, embedding the region token in backbone will disrupt the pre-trained model \cite{transvg++}, and the query anchor also depends on the decoder \cite{kamath2021mdetr}. With the unified feature space established by modality-shared encoder, we no longer need additional cross-modal en-/decoders to fuse uni-modal features, enabling us to more effectively leverage the knowledge acquired by pre-trained backbone. Benefiting from MRefM paradigm, the visual token inherently contains referring information. Consequently, we can discard special grounding token/anchors and directly construct lightweight and highly concise grounding and segmentation task heads based on the dot product operation within Referring MIM to unify the referring framework.

\textit{\textbf{Contributions:}} Our contributions are threefold: 
\textit{\textbf{(i)}} We pioneer the application of mask modeling to referring tasks by introducing a novel paradigm called mask referring modeling. This paradigm effectively models the referential relation between visual and language.
\textit{\textbf{(ii)}} Diverging from previous works, we propose a remarkably concise one-tower framework for grounding and referring segmentation in a unified modality-shared feature space. Our model eliminates the commonly used modality interaction modules, modality fusion en-/decoders, and special grounding tokens.
\textit{\textbf{(iii)}} We extensively validate the effectiveness of MRefM in three referring tasks on five datasets. Our method consistently surpasses existing approaches and achieves SoTA performance across several settings, providing a valuable new insights for future grounding and referring segmentation research.

\vspace{-3pt}
\section{Related work}
\label{2.0-relatedWork}
\vspace{-3pt}

\subsection{Referring expression comprehension (REC) and segmentation (RES)}
\label{2.1-rec}

\textit{\textbf{(i) REC.}} The recent supervised REC task, also known as visual grounding in a narrow sense, can be broadly categorized into \textbf{five main approaches}: 
\textbf{(1)} Fine-tuning with a uni-modal pre-trained language model and a closed-set detector. This setting is exemplified by TransVG \cite{deng2021transvg}, which builds upon the two-stage \cite{yu2018mattnet,cm-att-erase,liu2019learning,rvg-tree} and one-stage \cite{resc,yang2019fast,realgin} methods from the CNN era. It is considered the most conventional and extensively studied approach. 
\textbf{(2)} Fine-tuning with a pre-trained uni-modal language model and an open-set detection model pre-trained on box-level datasets mixed with multiple data sources. MDETR \cite{kamath2021mdetr} represents this type of setting, where \cref{fig:fig1}-(a)-(ii) plays a dominant role in its model structure. 
\textbf{(3)} Fine-tuning with multimodal self-supervised pre-trained models. CLIP-VG \cite{xiao2023clip} serves as an example for this category, introduced primarily through the proposal of CLIP \cite{radford2021learning}.
\textbf{(4)} Multimodal and multi-task mix-supervised pre-trained models. These methods typically combine multiple tasks while mixing datasets from each downstream task, employing mixed pre-training that incorporates both self-supervision and fine-grained supervision. UniTAB \cite{yang2022unitab}, OFA\cite{wang2022ofa}, \etc, represent such approaches where visual grounding often acts as one of the pre-training tasks.
\textbf{(5)} Grounding multimodal large language models (GMLLMs). These methods influenced by works like GPT \cite{gpt3} or LLAMA \cite{touvron2023llama} \etc These models integrate visual backbones into Large Language Models (LLMs) to generate grounding results rather than relying on regression techniques. Our approach mainly falls under type (3).
\textit{\textbf{(ii) RES.}} The development and approach categories of RES \cite{li2021mail, chng2023mask, kim2023risclip, wang2022cris, vg-law, lavt, yan2023universal, wang2024hierarchical} are generally similar to those of REC. However, the key distinction lies in the finer granularity of RES's output, which necessitates separate study from REC. In terms of model architecture, RES works predominantly employ two modality-dependent encoders and a decoder to generate the segmentation mask. Our work stands out as the first endeavor to explore RES within a unified multimodal feature space under a one-tower structure.



\vspace{-2pt}
\subsection{Mask vision language modeling}
\label{2.3-mvlm}
\vspace{-2pt}

Motivated by the success of MLM \cite{transformer} in BERT \cite{bert}, MAE \cite{he2022masked} and BEiT \cite{bao2021beit} have primary shifted their attention to MIM \cite{fang2023eva, wang2023masked, bachmann2022multimae}. Subsequently, exemplified by BEiT-3 \cite{beit3}, numerous MVLM works \cite{luo2022conditioned, li2023scaling, kim2023magvlt, zhao2022mamo, arici2021mlim} have emerged, with most of these works implementing randomly alternating uni-modal MIM and MLM. Most relevant to our work are mask region modeling (known as MRM) \cite{nguyen2023r-mae, wang2023masked}, which can be either unimodal MIM (\eg, R-MAE \cite{nguyen2023r-mae}) or employ more fine-grained regional data and contrastive learning to reconstruct the alignment between regions and object labels (\eg, ConLIP \cite{luo2022conditioned}, VLT \cite{ding2021vlt, ding2022vlt} \etc). However, our work focuses on modeling the fine-grained referential relationship within image and text, so as to enhance the cross-modal referring capability, which is significantly different from these works.

\begin{figure*}[t]
 \centering
   \includegraphics[width=1.0\linewidth]{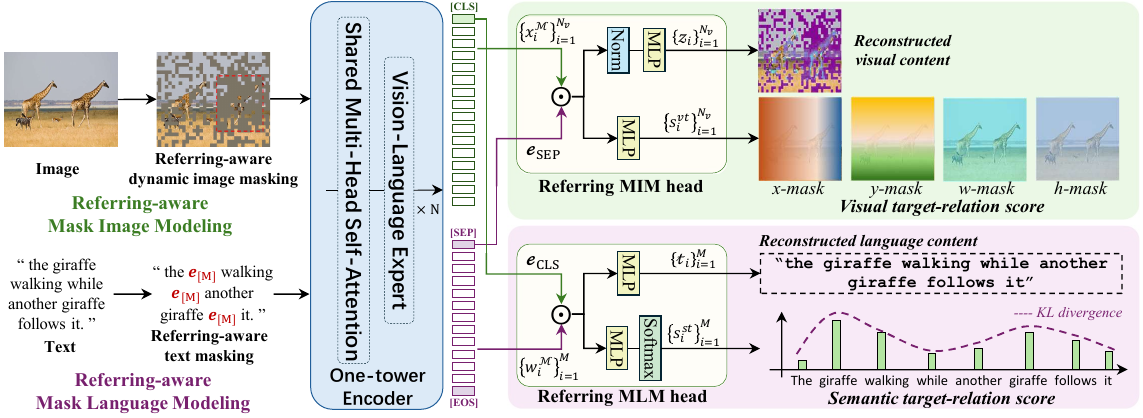}
   \vspace{-15pt}
   \caption{Illustration of our multimodal Mask Referring Modeling (MRefM) paradigm, which includes Referring-aware mask image modeling and Referring-aware mask language modeling.}
    \vspace{-7pt}
   \label{fig:fig2}
\end{figure*}

\section{Methodology}
\label{3.0-method}
\vspace{-2pt}

In this section, we propose our multimodal Mask Referring Modeling (\textit{\textbf{MRefM}}) paradigm, which includes Referring MIM and Referring MLM, as well as a feature space unified grounding and segmentation framework \textit{\textbf{OneRef}}. We will introduce these methods in the following sections.

Following BEiT-3 \cite{beit3}, we employ a multimodal modality-shared Transformer \cite{bao2022vlmo} as the underlying backbone network. Initially, we perform mask-then-predict MRefM pre-training, and followed by transfer fine-tuning on the referring tasks. As shown in \cref{fig:fig2}, the MRefM pre-training stage consists of two components: Referring-aware Mask Image Modeling (\textbf{Referring MIM}) and Referring-aware Mask Language Modeling (\textbf{Referring MLM}). Both modules aim to reconstruct two types of content: modality-related content within each modality and cross-modal fine-grained referring content.

\subsection{Preliminaries}
\label{3.1-preliminary}

BEiT-3 \cite{beit3} utilizes MIM, MLM, and MVLM for processing image, text, and image-text pairs respectively to facilitate the acquisition of general representations through MoE heads and shared multi-head self-attention. Notably, MVLM involves alternate training of MIM and MLM. Specifically:

\textbf{(i) Vanilla mask image modeling.}  We denote $\bm{x} \in \mathbb{R}^{H \times W \times 3}$ as the input image, and it is tokenized by a convolution projection to $N_v={HW}/{P^2}$ patches $\{\bm{x}^{p}_{i} \}_{i=1}^{N_v}$, where $\bm{x}^{p} \in \mathbb{R}^{N_v \times D}$, $H,W$ are the image size,  and $P$ is the patch size, $D$ is the hidden dimension of the unified feature space. 
Then, we leverage a specific masking strategy to mask a specific number of image patches. The masked position is termed as $\gM_v$. Thus, a shared learnable embedding $\ve_{[\text{M}]}$ is used to replace the masked image patch embeddings $\vx^p_{i}$ if $i \in \gM_v$. Subsequently, we prepend a learnable \texttt{[CLS]} token to the input, \ie, $[\ve_{\texttt{CLS}}, \{ \vx_{i}^{p} \}_{i=1}^{N_v}]$, and feed them to the one-tower Transformer. Next, we utilize a MIM head which consists of a linear projection and a softmax classifier to predict the visual tokens of the masked positions based on the corrupted image $\vx^{\gM}$. The visual tokens are obtained by the image tokenizer VQ-KD$_{\text{CLIP}}$ proposed in BEiT v2 \cite{Peng2022BEiTv2}, which provides supervisions for the MIM self-supervised learning procedure. The visual tokens of the original image are denote as $\{\vz_i\}_{i=1}^{N_v}$, and $\gI$ denotes the pre-training images. Then, the training loss of MIM is defined as:
\begin{align}
\small
\mathcal{L}_{\rm{MIM}} = - \sum_{\vx \in \gI} \sum_{i \in \gM_v} \mathrm{log} \ p( \vz_i | \vx^{\gM}_i ).
\label{eq:mim_objective}
\end{align}

\textbf{(ii) Vanilla mask language modeling.} The input text is tokenized and projected to the word embeddings ${\{\bm{w}_i\}}_{i=1}^{M}$ by a SentencePiece tokenizer \cite{kudo2018sentencepiece} with vocabulary size of 64010, where $\bm{w} \in \mathbb{R}^{M \times D}$, $M$ is the length of tokenized text sequence. Then, following BEiT-3 \cite{beit3}, we randomly mask the text tokens with a fixed masking ratio $\delta$. The masked position is termed as $\gM_w$.  Thus, a shared learnable embedding $\vw_{[\text{M}]}$ is used to replace the masked word tokens $\vw_{i}$ if $i \in \gM_w$. We prepend a learnable special tokens \texttt{[SEP]} and an end-of-sequence token \texttt{[EOS]} to the sequence, \ie, $[\ve_{\texttt{SEP}}, \{ \vw_{i}\}_{i=1}^{M}, \ve_\texttt{EOS}]$, and feed them to the one-tower Transformer. Similarly, we utilize a MLM head which consists of a linear projection to predict the text tokens of masked positions based on the corrupted text data $\vw^{\gM}$. The original textual tokens are denoted as $\{\vt_i\}_{i=1}^{M}$, and $\gT$ denotes the pre-training text sequences. Then, the training loss of MLM is defined as:
\vspace{3pt}
\begin{align}
\small
\mathcal{L}_{\rm{MLM}} = - \sum_{\vx \in \gT} \sum_{i \in \gM_w} \mathrm{log} \ p( \vt_i | \vw^{\gM}_i ).
\label{eq:mlm_objective}
\end{align}

\subsection{Referring-aware mask image modeling}
\label{subsec:ref_mim}

After concatenating the visual and text tokens and feeding them into the modality-shared encoder, the vanilla MVLM is commonly implemented through the alternating use of MIM and MLM \cite{beit3}. Despite the multimodal features are interact within the modality-shared encoder, it fundamentally remains a unimodal information reconstruction. Additionally, MVLM acquires general knowledge by randomly masking images and texts, it fails to effectively model the referential relationship. Hence, we propose Referring MIM and Referring MLM methods. Specifically, as shown in \cref{fig:fig2}, our proposed Referring MIM incorporates two additional components: the reconstruction of visual target-relation score and a referring-aware dynamic masking strategy.

In Referring MIM (\cref{fig:fig2}), instead of using uni-modal visual tokens \cite{beit3, li2023scaling, arici2021mlim}, we propose to employ visual tokens that dot product with the aggregated text token $\bm{e}_{\texttt{SEP}} \in \mathbb{R}^{1 \times D}$ for the reconstruction purpose. The reconstruction of Referring MIM involves not only the modality-related content $\{\vz_{i}\}_{i\in \gM_v}$ but also the visual target-relation scores $\{\vs_{i}^{vt} \}_{i=1}^{N_v} \in \mathbb{R}^{N_v \times 4}$. We utilize a visual target-relation head which consists of a three-layer perceptron (MLP) to predict the scores. The scores represent the distance between each patch token $\{\vx_{i}^{\gM} \}_{i=1}^{N_v}$ and the referred region $\mathcal{B}=(x_c,y_c,w_r,h_r)$, where $(x_c,y_c,w_r,h_r)$ denote the center coordinate and the width and height of the referred region. It encompasses four masks, \ie, \textit{x-, y-, w-, h- masks}, which represent the normalized horizontal and vertical distances from the referred center, \ie, $((x-x_c)/W,\ (y-y_c)/H)$, and the proportion of width and height on the the referred region, \ie, $(P/w_r,\ P/h_r)$, respectively, where $(x,y)$ denote the center coordinate of each patch. We denote $\odot$ as dot product operation. Finally, the training loss of Referring MIM is defined as:
\begin{equation}
\small
\mathcal{L}_{\rm{Referring~MIM}} = - \sum_{\vx \in \gI} \sum_{i \in \gM_v} \mathrm{log} \ p( \vz_i | (\vx^{\gM}_i \odot \bm{e}_{\texttt{SEP}}))
- \sum_{\vx \in \gI} \sum_{i \in [1,N_v]} \mathrm{log} \ p( \vs_i^{vt} | (\vx^{\gM}_i \odot \bm{e}_{\texttt{SEP}})).
\label{eq:ref_mim}
\end{equation}

\begin{wrapfigure}{R}{0.54\textwidth}
\vspace{-24pt}
\begin{minipage}{0.54\textwidth}
\begin{algorithm}[H]
    \small
    \SetAlgoLined
    \caption{Referring-aware Dynamic Masking}
    \label{alg:mask}
    \KwIn{$N_v$ \textit{image patches}, $N_r$ ($h_{rp}\times w_{rp}$) \textit{referred patches}.} 
    \KwOut{\textit{Dynamic masked positions} $\mathcal{M}$.} 
    $~~~\bm{c} \gets$ \textsf{Rand Select} $\beta \cdot N_v$ numbers in $[1, N_v]$      \\
    \textsf{New} $\mathcal{M}\in \mathbb{R}^{1 \times N_v}$, $\{$$\{\mathcal{M}_i\}_{i}^{N_v} \mid \mathcal{M}_i = 1$ if $i \in \bm{c}$, else 0$\}$   \\
    $\mathcal{M} \gets$ $\mathcal{M}$ reshape as $\mathcal{M} \in \mathbb{R}^{h \times w}$  
    \hfill\AlgCommentInLine{In-context masking} \\
    \textsf{New} $\mathcal{M}_r \in \mathbb{R}^{h_{rp} \times h_{rp}}$ with all as $0$  
    \hfill\AlgCommentInLine{Referred masking}\\
    \While{$|\mathcal{M}_r| \leq \gamma \cdot N_r$}{  
        $~~s \gets$ \textsf{Rand(}$1, \gamma \cdot N_r - |\mathcal{M}_r|$\textsf{)}
        \hfill\AlgCommentInLine{Block size}  \\
        $r \gets$ \textsf{Rand($a$, $\frac{1}{a}$\textsf{)}}  \hfill\AlgCommentInLine{Aspect ratio of block}  \\
        $w_b \gets \sqrt{s/r}; h_b \gets \sqrt{s \cdot r}$ \hfill\AlgCommentInLine{Width, height of block}  \\
        $l \gets$ \textsf{Rand($0$, $w_{rp} - w_b$\textsf{)}}; $t \gets$ \textsf{Rand($0$, $h_{rp} - h_b$\textsf{)}}  \\
        $\{\mathcal{M}_r(i,j)=1 \mid i \in [l,l+w_b), j \in [t,t+h_b) \}$   
    } 
    $~~\mathcal{M}(x_{sp}:x_{sp}+w_{rp},y_{sp}:y_{sp}+h_{rp}) = \mathcal{M}_r$   \\
    \Return $\mathcal{M}$.
\end{algorithm}
\end{minipage}
\vspace{-14pt}
\end{wrapfigure}

\textbf{Referring-aware dynamic image masking strategy.} As shown in \cref{fig:mask_sample}, among the existing masking strategies, MAE \cite{he2022masked} adopts a high-ratio random masking while BEiT-3 \cite{beit3} uses a low-ratio block-wise random masking, neither of which effectively directs attention to the referred region. SemMAE \cite{li2022semmae} proposes a semantic-guided masking that requires additional bulky semantic models and limits its generality. To enhance the model's understanding of the referred region through surrounding visual context and text semantics, we propose a referring-aware dynamic masking strategy as shown in \cref{alg:mask}.

The strategy avoids the drawbacks of the aforementioned methods and directs the model's attention to the referred region. Specifically, we denote the shape after patch reshaping of the image as $(h,w)$, where $h=H/P$, $w=W/P$, and $N_v = h\times w$. To maximize the masking of the referred region $(x_s,y_s,w_r,h_r)$, where $x_s$, $y_s$ represent the starting coordinates of the referred region, we introduce a margin $m$ to its surroundings and denote its patch coordinates as $(x_{sp},y_{sp},w_{rp},h_{rp})$, \ie, $x_{sp}=\lfloor{x_s/P}\rfloor-m$, $w_{rp}=\lfloor{w_r/P}\rfloor+m$, $y_{sp}$ and $h_{rp}$ are similar to that of $x_{sp}$ and $w_{rp}$, where $\lfloor \cdot \rfloor$ indicates rounding down to an integer. Thus, the number of referred patches is denote as $N_r=h_{rp}\times w_{rp}$. Then, as shown in \cref{alg:mask}, to ensure that the model allocates appropriate attention to the in-contextual information around the referred region, we utilize a random masking with a relatively low ratio $\beta$ for its surroundings. Simultaneously, we employ a block-wise masking approach with a high ratio $\gamma$ in the extended area of the region. Since referred regions vary across different image-text pairs, each sample's entire masking ratio $\alpha$ is dynamically determined:
\begin{align}
\alpha = [\beta \cdot (N_v - N_r) + \gamma \cdot N_r] / N_v.
\label{eq:mask_ratio}
\end{align}

\begin{figure*}[t]
\vspace{-17pt}
 \centering
   \includegraphics[width=1.0\linewidth]{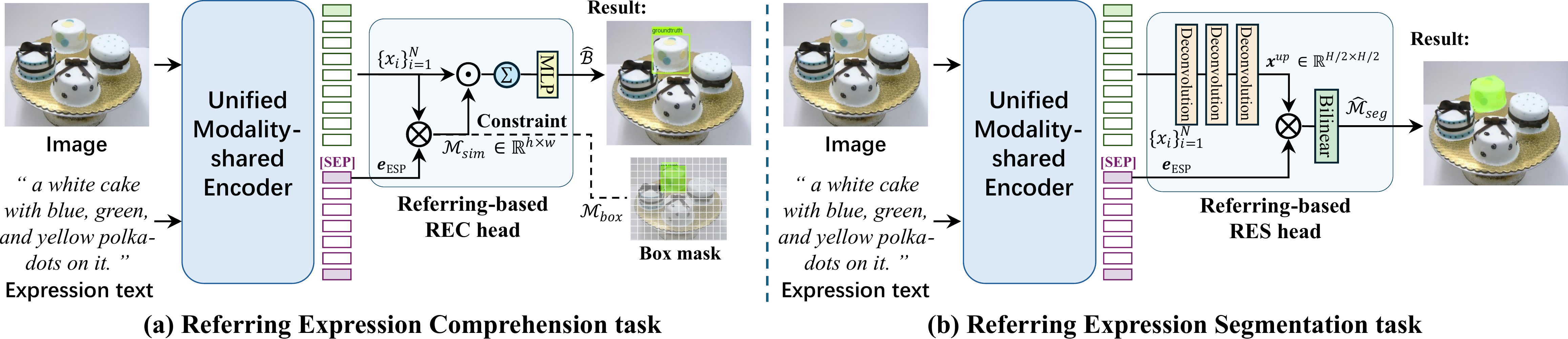}
   \vspace{-16pt}
   \caption{Illustration of the referring-based grounding and segmentation transfer.}
    \vspace{-10pt}
   \label{fig:fig3}
\end{figure*}

\subsection{Referring-aware mask language modeling}
\label{subsec:ref_mlm}


Similarly, in Referring MLM, instead of using uni-modal linguistic tokens \cite{beit3, li2023scaling, arici2021mlim}, we propose to employ linguistic tokens that dot product with the aggregated visual token $\bm{e}_{\texttt{CLS}} \in \mathbb{R}^{1 \times D}$ for the reconstruction purpose. The reconstruction of Referring MLM involves not only the modality-related content $\{\vt_{i}\}_{i\in \gM_w}$ but also the semantic target-relation scores $\{\vs_{i}^{st} \}_{i=1}^{M}$. The score represents the correlation between the referred target and the language token, which is obtained by a teacher model (\ie, a BEiT-3 model with performed image-text contrastive intermediate tuning) with calculating the weighted sum of the normalized similarity between the language token $\{\vw_{i}\}_{i=1}^{M}$ and the aggregated visual token $\bm{e}_{\texttt{CLS}}^{reg}$ of referred region, as well as the aggregated visual token $\bm{e}_{\texttt{CLS}}^{img}$ of entire image:
\begin{equation}
\small
\vs^{st}=\lambda_{reg} \cdot \sigma(<{\bm{e}_{\texttt{CLS}}^{reg}}^\top,\ \{\vw_{i}\}_{i=1}^{M}>)
+ \lambda_{img} \cdot \sigma(<{\bm{e}_{\texttt{CLS}}^{img}}^\top,\ \{\vw_{i}\}_{i=1}^{M}>),
\label{eq:sem_target}
\end{equation}
where $<\cdot, \cdot>$ denotes cosine similarity operation, $\sigma$ denotes the softmax normalization.As shown in \cref{fig:fig2}, we utilize a semantic target-relation head which consists of a three-layer MLPs and a softmax normalization to predict the scores. Finally, the training loss of Referring MLM is defined as:
\begin{equation}
\small
\mathcal{L}_{\rm{Referring~MLM}} = - \sum_{\vw \in \gT} \sum_{i \in \gM_w} \mathrm{log} \ p( \vt_i | (\vw^{\gM}_i \odot \bm{e}_{\texttt{CLS}}))
- \sum_{\vw \in \gT} \sum_{i \in [1,M]} \mathrm{log} \ {p_{kl}}( \vs_i^{st} | (\vw^{\gM}_i \odot \bm{e}_{\texttt{CLS}})),
\label{eq:ref_mlm}
\end{equation}
where ${p_{kl}}$ represents a probabilistic prediction with Kullback-Leibler divergence \cite{hinton2015distilling}.

\vspace{-5pt}
\subsection{Referring-based grounding and segmentation transfer}
\vspace{-4pt}

The modeling of visual and language in a unified feature space eliminates the need for the commonly-used Transformer-based fusion en-/decoder \cite{deng2021transvg, liu2023dqdetr, liu2023grounding} and various early-stage interaction techniques \cite{transvg++, vg-law, xiao2024hivg} to further uniform the visual and language features. Additionally, since the referential relationship is modeled by MRefM during pre-training, we can accurately regress the results of grounding and referring segmentation using the output tokens, without relying on the widely-used special grounding tokens (\eg, \texttt{[Region]} token \cite{deng2021transvg, transvg++, qrnet, xiao2023clip, xiao2024hivg}, query anchors \cite{liu2023grounding, liu2023dqdetr}).

\textbf{Referring expression comprehension.} As illustrated in \cref{fig:fig3}-(a), based on Referring MIM, we initially perform a similarity operation between visual tokens $\{\vx_{i}\}_{i=1}^{N_v} \in \mathbb{R}^{N_v \times D}$ and aggregated language token $\bm{e}_{\texttt{SEP}} \in \mathbb{R}^{1 \times D}$ to obtain a softmax-normalized similarity mask $\mathcal{M}_{sim} \in \mathbb{R}^{h \times w}$. This mask is then replicated and multiplied back to each hidden dimension of the visual tokens. Subsequently, the visual tokens are summed to yield reduced tokens, which are finally subjected to regress the prediction box $\hat{\mathcal{B}}=(\hat{x}_c,\hat{y}_c,\hat{w}_r,\hat{h}_r)$ using a 3-layer MLPs:
\begin{equation}
\small
\hat{\mathcal{B}} = {\rm{MLP}}(\sum_{i \in [1,N_v]}({\rm{Repeat}}( \sigma(<{\bm{e}_{\text{ESP}}}^\top,\ \{\vx_{i}\}_{i=1}^{N_v}>)) \odot {\rm{MLP}}(\{\vx_{i}\}_{i=1}^{N_v}))). 
\end{equation}

\vspace{-10pt}
To enhance the accuracy of cross-modal similarity, we propose treating the similarity as a coarse-grained downsampling bounding box mask $\mathcal{M}_{box} \in \mathbb{R}^{h \times w}$ and imposing segmentation loss (\ie, Focal loss \cite{lin2017focal} and Dice/F-1 loss \cite{milletari2016v}) on the sigmoid activated similarity mask $\mathcal{M}_{sim} \in \mathbb{R}^{h \times w}$ with coefficient $\lambda_{f\_box}$ and $\lambda_{d\_box}$ as the box mask constraints: 
\begin{equation}
\mathcal{L}_{box\_mask\_constraints} = \lambda_{f\_box}\mathcal{L}_{focal}(\mathscr{M}_{sim},\mathcal{M}_{box}) + \lambda_{d\_box}\mathcal{L}_{dice}(\mathcal{M}_{sim},\mathcal{M}_{box}).
\end{equation}

Consequently, the loss function for the REC task can be reformulated as the weighted sum of vanilla grounding loss (\ie, smooth L1 loss \cite{girshick2015fast} and Giou loss \cite{rezatofighi2019generalized}) and box mask constraints:
\begin{equation}
\mathcal{L}_{REC}={\lambda_{L_1}}\mathcal{L}_{\text{L1}}\big(\hat{\mathcal{B}},\mathcal{B}\big)+{\lambda_{giou}}\mathcal{L}_{\text{giou}}\big(\hat{\mathcal{B}},\mathcal{B}\big) + \mathcal{L}_{box\_mask\_constraints}.
\label{eq:loss1}
\end{equation}

\vspace{-5pt}
\textbf{Referring expression segmentation.} As illustrated in \cref{fig:fig3}-(b), the implementation of referring segmentation can be regarded as a simplified version of grounding. Initially, we employ a 3-layer deconvolution to up-sample the visual token to $\vx^{up} \in \mathbb{R}^{H/2 \times W/2}$. Subsequently, cosine similarity operations are performed on the up-sampled visual tokens and the aggregated language token. The resulting similarity mask is then utilized as the final predicted mask $\hat{\mathcal{M}}_{seg} \in \mathbb{R}^{H \times W}$ after applying 1-layer bilinear interpolation. We denote the ground truth segmentation mask as $\mathcal{M}_{seg} \in \mathbb{R}^{H \times W}$, then the loss function for RES is defined as follows:
\begin{equation}
\mathcal{L}_{RES} = \lambda_{f\_seg}\mathcal{L}_{focal}(\hat{\mathcal{M}}_{seg},\mathcal{M}_{seg}) + \lambda_{d\_seg}\mathcal{L}_{dice}(\hat{\mathcal{M}}_{seg},\mathcal{M}_{seg}).
\end{equation}

\begin{table*}[t]
\vspace{-7pt}
    \footnotesize
    \caption{Comparison with \textbf{latest} SoTA methods on the five datasets for \textbf{REC/PG} tasks with single-dataset fine-tuning setting. We highlight best result of base model in \textcolor{red}{red} and \textbf{bold} for large model.}
    \centering
    \resizebox{1.0\columnwidth}{!}{%
    \begin{tabular}{c|c|c|c|ccc|ccc|cc|c|c}
    \toprule
    \multirow{2}[2]{*}{Methods} & \multirow{2}[2]{*}{Venue} & Visual   & Language  & \multicolumn{3}{c|}{RefCOCO} & \multicolumn{3}{c|}{RefCOCO+} & \multicolumn{2}{c|}{RefCOCOg} & ReferIt & Flickr \\
                                &                           & Backbone & Backbone  &   val  & testA & testB       &   val  & testA & testB        &   val  & test                 & test    & test   \\
    \midrule    
    \multicolumn{14}{l}{\textbf{Single-dataset fine-tuning setting \textit{w.} uni-modal pre-trained close-set detector and language model: (traditional setting)}}        \\
    TransVG \cite{deng2021transvg}      &   ICCV'21     & RN101+DETR  & BERT-B    & 81.02  & 82.72  & 78.35  & 64.82  & 70.70  & 56.94  & 68.67  & 67.73  & 70.73 & 79.10  \\
    Word2Pix \cite{zhao2022word2pix}    &   TNNLS'22    & RN101+DETR  & BERT-B    & 81.20  & 84.39  & 78.12  & 69.74  & 76.11  & 61.24  & 70.81  & 71.34  & --    & --     \\
    QRNet \cite{qrnet}          &   CVPR'22     & Swin-S \cite{liu2021swin} & BERT-B     & 84.01  & 85.85  & 82.34  & 72.94  & 76.17  & 63.81  & 71.89  & 73.03  & 74.61 & 81.95 \\   
    VG-LAW \cite{vg-law}        &   CVPR'23     & ViT-Det \cite{vit-det} & BERT-B     & 86.06  & 88.56  & 82.87  & 75.74  & 80.32  & 66.69  & 75.31  & 75.95  & 76.60 & --    \\
    TransVG++\cite{transvg++}   &   TPAMI'23    & ViT-Det \cite{vit-det} & BERT-B     & 86.28  & 88.37  & 80.97  & 75.39  & 80.45  & 66.28  & 76.18  & 76.30  & 74.70 & 81.49 \\ 
    
    \midrule  
    \multicolumn{14}{l}{\textbf{Single-dataset fine-tuning setting \textit{w.} vision-language self-supervised pre-trained model: }}       \\ 
    CLIP-VG \cite{xiao2023clip} &   TMM'23      & CLIP-B & CLIP-B     & 84.29  & 87.76  & 78.43  & 69.55  & 77.33  & 57.62  & 73.18  & 72.54  & 70.89 & 81.99 \\  
    JMRI \cite{zhu2023jmri} &   TIM'23      & CLIP-B & CLIP-B     & 82.97 & 87.30  & 74.62  & 71.17  & 79.82  & 57.01  & 71.96  & 72.04  & 68.23 & 79.90 \\  
    Dynamic-MDETR & TPAMI'23    & CLIP-B & CLIP-B     & 85.97  & 88.82  & 80.12  & 74.83  & 81.70  & 63.44  & 74.14  & 74.49  & 70.37 & 81.89 \\
    HiVG-B \cite{xiao2024hivg}  &   ACMMM'24      & CLIP-B & CLIP-B       &   87.32    &   89.86    &   83.27    &   78.06    &   83.81    &   68.11    &   78.29    &   78.79    &  75.22  &  82.11 \\
    HiVG-L \cite{xiao2024hivg}  &   ACMMM'24     & CLIP-L & CLIP-L       &  88.14  & 91.09  & 83.71  & 80.10  &  86.77  & 70.53  & 80.78  & 80.25  & 76.23 & 82.16 \\
    \midrule 
    \rowcolor{cyan!03}
    \textbf{OneRef-B (ours)}      &   NeurIPS'24     & BEiT3-B & BEiT3-B       &   \textcolor{red}{88.75}    &   \textcolor{red}{90.95}    &   \textcolor{red}{85.34}    &   \textcolor{red}{80.43}    &   \textcolor{red}{86.46}    &   \textcolor{red}{74.26}    &   \textcolor{red}{83.68}    &   \textcolor{red}{83.52}    &  \textcolor{red}{77.17}  &  \textcolor{red}{83.61} \\ \rowcolor{cyan!03}
    \textbf{OneRef-L (ours)}      &   NeurIPS'24     & BEiT3-L & BEiT3-L       & \textbf{92.87}  & \textbf{94.01}  & \textbf{90.19}  & \textbf{87.98}  & \textbf{91.57}  & \textbf{83.73}  & \textbf{88.11}  & \textbf{89.29}  & \textbf{81.11} & \textbf{84.75} \\
    \bottomrule
    \end{tabular}%
    }
    \label{tab:rec_sota1}%
  \end{table*}%
    

\begin{table*}[t]
\vspace{-5pt}
    \footnotesize
    \caption{Comparison with \textbf{latest} SoTA methods for \textbf{REC} task with dataset-mixed intermediate pre-training setting. `RefC' represents the mixup of RefCOCO/+/g training data. $\dagger$ indicates RefC has been used during pre-training. `G-DINO-L$^*$' denotes `O365,OI,GoldG,Cap4M,COCO,RefC'.}
    \centering
    \resizebox{1.0\columnwidth}{!}{%
    \begin{tabular}{c|c|c|c|c|ccc|ccc|cc}
    \toprule
    \multirow{2}[2]{*}{Methods} & \multirow{2}[2]{*}{Venue} & Visual/Language & Intermediate  & Data & \multicolumn{3}{c|}{RefCOCO} & \multicolumn{3}{c|}{RefCOCO+} & \multicolumn{2}{c}{RefCOCOg}  \\
                                &                           & Backbone        & pretrain data & size   &   val  & testA & testB       &   val  & testA & testB        &   val  & test                  \\
    \midrule       
    \multicolumn{13}{l}{\textbf{Dataset-mixed intermediate pre-training setting (\textit{w.} box-level dataset-mixed open-set detection pre-trained model)}}       \\ 
     MDETR $^\dagger$ \cite{kamath2021mdetr}     & ICCV'21   & RN101/RoBERT-B        & GoldG,RefC          & 6.5M  & 86.75  & 89.58  & 81.41  & 79.52  & 84.09  & 70.62  & 81.64  & 80.89  \\  
     YORO$^\dagger$ \cite{ho2023yoro}  & ECCV'22 & ViLT \cite{kim2021vilt} / BERT-B  & GoldG,RefC          & 6.5M  & 82.90  & 85.60  & 77.40  & 73.50  & 78.60  & 64.90  & 73.40  & 74.30  \\
     DQ-DETR $^\dagger$ \cite{liu2023dqdetr}     & AAAI'23   & RN101 / BERT-B        & GoldG,RefC          & 6.5M  & 88.63  & 91.04  & 83.51  & 81.66  & 86.15  & 73.21  & 82.76  & 83.44  \\
     Grounding-DINO-B$^\dagger$                  & arXiv'23  & Swin-T     / BERT-B   & O365,GoldG,RefC     & 7.2M  & 89.19  & 91.86  & 85.99  & 81.09  & 87.40  & 74.71  & 84.15  & 84.94  \\
     Grounding-DINO-L$^\dagger$                  & arXiv'23  & Swin-L     / BERT-B   & G-DINO-L$^*$     & 21.4M & 90.56  & 93.19  & 88.24  & 82.75  & 88.95  & 75.92  & 86.13  & 87.02  \\
     CyCo \cite{wang2024cycle}                   & AAAI'24   & ViT\cite{vit}/ BERT-B & VG,SBU,CC3M,\etc    & >120M & 89.47  & 91.87  & 85.33  & 80.40  & 87.07  & 69.87  & 81.31  & 81.04  \\
     HiVG-B$^\dagger$ \cite{xiao2024hivg}        & ACMMM'24  & CLIP-B / CLIP-B       & RefC,ReferIt,Flickr & 0.8M  & 90.56  & 92.55  & 87.23  & 83.08  & 87.83  & 76.68  & 84.71  & 84.69  \\
     HiVG-L$^\dagger$ \cite{xiao2024hivg}        & ACMMM'24  & CLIP-L / CLIP-L       & RefC,ReferIt,Flickr & 0.8M  & 91.37  & 93.64  & 88.03  & 83.63  & 88.16  & 77.37  & 86.73  & 86.86  \\
    \midrule       
    \multicolumn{13}{l}{\textbf{Fine-tuning setting \textit{w.} dataset-mixed multi-task mix-supervised pre-trained model: }}       \\ 
     UniTAB $^\dagger$ \cite{yang2022unitab}     & ECCV'22   & RN101/RoBERT-B & VG,COCO,\etc & >20M  & 88.59  & 91.06 & 83.75  & 80.97  & 85.36  & 71.55  & 84.58  & 84.70   \\
     OFA-B $^\dagger$ \cite{wang2022ofa}         & ICML'22   & OFA-B      / OFA-B    & -- & --  & 88.48  & 90.67  & 83.30  & 81.39  & 87.15  & 74.29  & 82.29  & 82.31   \\
     OFA-L $^\dagger$ \cite{wang2022ofa}         & ICML'22   & OFA-L      / OFA-L    & -- & --  & 90.05  & 92.93  & 85.26  & 85.80  & 89.87  & 79.22  & 85.89  & 86.55   \\
        \midrule  
    \multicolumn{13}{l}{\textbf{Fine-tuning setting \textit{w.} grounding multimodal large language model (GMLLM): }}       \\ 
     Shikra-7B$^\dagger$ \cite{chen2023shikra}  &   arXiv'23     & CLIP-L / Vicuna-7B\cite{chiang2023vicuna} & RefC,VG &  0.5M   & 87.01 & 90.61 & 80.24 & 81.60 & 87.36 & 72.12 & 82.27 & 82.19  \\
     Ferret-7B$^\dagger$ \cite{you2023ferret}  &   ICLR'24     & CLIP-L / Vicuna-7B\cite{chiang2023vicuna} & GRIT \cite{you2023ferret} & >8M    & 87.49 & 91.35 & 82.45 & 80.78 & 87.38 & 73.14 & 83.93 & 84.76   \\
     LION-4B$^\dagger$ \cite{chen2024lion}  &   CVPR'24     & EVA-G\cite{fang2023eva}/FlanT5-3B & VG,COCO,\etc & 3.6M    & 89.73 & 92.29 & 84.82 & 83.60 & 88.72 & 77.34 & 85.69 & 85.63  \\
     LION-12B$^\dagger$ \cite{chen2024lion} &   CVPR'24     & EVA-G\cite{fang2023eva}/FlanT5-11B & VG,COCO,\etc & 3.6M   & 89.80 & 93.02 & 85.57 & 83.95 & 89.22 & 78.06 & 85.52 & 85.74  \\
        \midrule
    \rowcolor{pink!06}
     \multicolumn{2}{l|}{~~\textbf{OneRef-B$^\dagger$ (unsupervised)}}           & BEiT3-B / BEiT3-B & RefC,ReferIt  & 0.5M     & 89.16 & 92.03  & 87.26 & 83.18 & 88.56 & 77.66  & 84.72 & 85.17  \\ 
    \rowcolor{cyan!03}
    \textbf{OneRef-B$^\dagger$ (0.2B)}        &   NeurIPS'24      & BEiT3-B / BEiT3-B & RefC,ReferIt & 0.5M     &   \textcolor{red}{91.89}    &   \textcolor{red}{94.31}    &   \textcolor{red}{88.58}    &   \textcolor{red}{86.38}    &   \textcolor{red}{90.38}    &   \textcolor{red}{79.47}    &   \textcolor{red}{86.82}    &   \textcolor{red}{87.32}      \\
    \rowcolor{cyan!03}
    \textbf{OneRef-L$^\dagger$ (0.6B)} &   NeurIPS'24   & BEiT3-L / BEiT3-L & RefC,ReferIt  & 0.5M   &  \textbf{93.21} & \textbf{95.43} & \textbf{90.11}  & \textbf{88.35}  & \textbf{92.11}  & \textbf{82.70}  &  \textbf{87.81} &  \textbf{88.83}  \\
    \bottomrule
    \end{tabular}%
    }
    \label{tab:rec_sota2}%
    \vspace{-13pt}    
\end{table*}%

\begin{table*}[t]
\vspace{-7pt}
    \footnotesize
    \caption{Comparison with \textbf{latest} SoTA methods (\textbf{mIoU} metric) on the three datasets for \textbf{RES} task with both single-dataset fine-tuning setting and dataset-mixed intermediate pre-training setting.}
    \centering
    \resizebox{1.0\columnwidth}{!}{%
    \begin{tabular}{c|c|c|c|ccc|ccc|cc}
    \toprule
    \multirow{2}[1]{*}{Methods} & \multirow{2}[1]{*}{Venue} & Visual/Language & Intermediate & \multicolumn{3}{c|}{RefCOCO} & \multicolumn{3}{c|}{RefCOCO+} & \multicolumn{2}{c}{RefCOCOg}  \\
                                &                           & Backbone        & pretrain data &   val  & testA & testB       &   val  & testA & testB        &   val  & test                  \\
    \midrule
    \multicolumn{12}{l}{\textbf{Single-dataset fine-tuning setting \textit{w.} uni-modal pre-trained close-set segmentation model: (traditional setting)}}        \\
    RefTR \cite{li2021referring}    &   NIPS'21  & RN101 / BERT-B    &  --    & 74.34  & 76.77 & 70.87 &  66.75 & 70.58 & 59.40 & 66.63 & 67.39  \\
    SeqTR \cite{zhu2022seqtr}       &   ECCV'22  & DN53\cite{redmon2018yolov3}/Bi-GRU   &  --    & 71.70  & 73.31 & 69.82 &  63.04 & 66.73 & 58.97 & 64.69 & 65.74 \\   
    LAVT \cite{lavt}                &   CVPR'22  & Swin-B / BERT-B   &  --    & 74.46  & 76.89 & 70.94 &  65.81 & 70.97 & 59.23 & 63.34 & 63.62  \\
    VG-LAW \cite{vg-law}            &   CVPR'23  & ViT-Det / BERT-B  &  --    & 75.05  & 77.36 & 71.69 &  66.61 & 70.30 & 58.14 & 65.36 & 65.13  \\
    \midrule  
    \multicolumn{12}{l}{\textbf{Single-dataset fine-tuning setting \textit{w.} vision-language self-supervised pre-trained model: }}       \\ 
    CRIS \cite{wang2022cris}        &   CVPR'22  & CLIP-L / CLIP-L &  --    & 70.47  & 73.18 & 66.10 &  62.27 & 68.06 & 53.68 & 59.87 & 60.36  \\
    JMCELN\cite{huang2023jmceln}    &  EMNLP'23  & CLIP-B / CLIP-B &  --    & 74.40  & 77.69 & 70.43 &  66.99 & 72.69 & 57.34 & 64.08 & 64.99  \\
    RISCLIP-B \cite{kim2023risclip} &  NAACL'24  & CLIP-B / CLIP-B &  --    & 75.68  & 78.01 & 72.46 &  69.16 & 73.53 & 60.68 & 67.62 & 67.97  \\
    RISCLIP-L \cite{kim2023risclip} &  NAACL'24  & CLIP-L / CLIP-L &  --    & 78.87  & 81.46 & 75.41 &  74.38 & 78.77 & 66.84 & 71.82 & 71.65  \\
    \midrule
    \rowcolor{cyan!03}
    \textbf{OneRef-B (ours)}        &   NeurIPS'24      & BEiT3-B / BEiT3-B & --      &   \textcolor{red}{77.57}    &   \textcolor{red}{79.05}    &   \textcolor{red}{75.11}    &   \textcolor{red}{71.25}    &   \textcolor{red}{75.41}    &   \textcolor{red}{65.45}    &   \textcolor{red}{69.37}    &   \textcolor{red}{69.70}      \\
    \rowcolor{cyan!03}
    \textbf{OneRef-L (ours)} &   NeurIPS'24   & BEiT3-L / BEiT3-L &  --    &  \textbf{80.09} & \textbf{82.19} & \textbf{77.51}  & \textbf{75.17}  & \textbf{79.38}  & \textbf{70.17}  &  \textbf{73.18} &  \textbf{73.76}  \\
    \midrule
    \midrule
    \multicolumn{12}{l}{\textbf{Dataset-mixed intermediate pre-training setting:}}       \\ 
    PolyFormer-B$^\dagger$\cite{liu2023polyformer} &  CVPR'23  & Swin-B / BERT-B &  RefC    & 75.96 & 77.09 & 73.22 & 70.65 & 74.51 & 64.64 & 69.36 & 69.88  \\
    RISCLIP-B$^\dagger$\cite{kim2023risclip}       &  NAACL'24 & CLIP-B / CLIP-B &  RefC    & 75.68 & 78.01 & 72.46 & 72.46 & 74.30 & 61.37 & 69.49 & 69.53  \\
    RISCLIP-L$^\dagger$\cite{kim2023risclip}       &  NAACL'24 & CLIP-L / CLIP-L &  RefC    & 79.53 & 81.78 & 75.78 & 74.88 & 78.88 & 68.09 & 73.45 & 74.52  \\
        \midrule  
    \rowcolor{pink!06}
     \multicolumn{2}{l|}{~~\textbf{OneRef-B$^\dagger$ (unsupervised)}}          & BEiT3-B / BEiT3-B & RefC  & 78.20 & 79.26  & 75.92 & 72.54 & 75.54 & 67.39 & 71.28  & 71.13  \\ 
    \rowcolor{cyan!03}
    \textbf{OneRef-B$^\dagger$ (ours)}        &   NeurIPS'24    & BEiT3-B / BEiT3-B & RefC      &   \textcolor{red}{79.83}    &   \textcolor{red}{81.86}    &   \textcolor{red}{76.99}    &   \textcolor{red}{74.68}    &   \textcolor{red}{77.90}    &   \textcolor{red}{69.58}    &   \textcolor{red}{74.06}    &   \textcolor{red}{74.92}      \\
    \rowcolor{cyan!03}
    \textbf{OneRef-L$^\dagger$ (ours)} &  NeurIPS'24   & BEiT3-L / BEiT3-L &  RefC    &  \textbf{81.26} & \textbf{83.06} & \textbf{79.45}  & \textbf{76.60}  & \textbf{80.16}  & \textbf{72.95}  &  \textbf{75.68} &  \textbf{76.82}  \\
    \bottomrule
    \end{tabular}%
    }
    \label{tab:res_sota1}%
    \vspace{-5pt}    
  \end{table*}%

\begin{table*}[t]
\centering
\begin{minipage}[t]{0.60\textwidth}
\centering
    \vspace{-5pt}
    \footnotesize
    \caption{Ablation of MRefM on mixup pre-training setting.}
    \centering
    \resizebox{1.0\columnwidth}{!}{%
    \begin{tabular}{c|c|c|ccc|cc}
    \toprule
    \multirow{2}[1]{*}{MIM} & \multirow{2}[1]{*}{MLM} & image masking & \multicolumn{3}{c|}{RefCOCO+} & \multicolumn{2}{c}{RefCOCOg}  \\
        &     & strategy   & val  & testA & testB        &   val  & test                 \\
    \midrule
    \XSolidBrush    &   \XSolidBrush  & \XSolidBrush       & 78.56  & 83.36 & 71.72 & 80.41 & 80.52  \\
    vanilla         &   vanilla       &  random            & 79.68  & 84.59 & 72.11 & 81.35 & 81.11  \\
    vanilla         &   vanilla       &  referring-aware   & 80.06  & 85.77 & 73.96 & 81.96 & 82.16  \\
    Ref MIM         &   vanilla       &  referring-aware   & 83.64  & 88.26 & 76.58 & 83.55 & 85.86  \\
    vanilla         &   Ref MLM       &  referring-aware   & 81.52  & 86.87 & 75.93 & 82.88 & 84.32  \\
    Ref MIM         &   Ref MLM       &  random            & 85.08  & 89.12 & 78.56 & 85.57 & 86.89  \\
    \rowcolor{cyan!03}
    Ref MIM         &   Ref MLM       &  referring-aware    &  \textbf{86.38}  & \textbf{90.38} & \textbf{79.47} & \textbf{86.82} & \textbf{87.32}  \\
    \bottomrule
    \end{tabular}%
    }
    \label{tab:ablation_mrm}%
    \vspace{-15pt}    
\end{minipage}
\hspace{1.5mm}
\begin{minipage}[t]{0.365\textwidth}
\centering
    \footnotesize
    \vspace{-4pt}
    \caption{Ablation of the task heads.}
    \vspace{-13pt}
    \begin{center}
    \resizebox{1.0\columnwidth}{!}{%
    \begin{tabular}{l|c|c}
        \toprule
        \multirow{2}[1]{*}{Architecture (Fine-tuning setting)}  &   \multicolumn{2}{c}{RefCOCOg}  \\
                   &   val  & test  \\
        \midrule
        \rowcolor{cyan!03}
        full model in REC                             & \textbf{83.68}   & \textbf{83.52}   \\
        full model \textit{w/o} box mask loss         & 82.54   & 82.02   \\
        \textit{w.} fusion encoder + region token     & 78.93   & 78.51   \\
        \midrule
        \rowcolor{cyan!03}
        full model in RES                    &  \textbf{69.37}  & \textbf{69.70}   \\
        deconv after similarity operation             &  67.98  & 68.62   \\
        4-layer deconv \textit{w/o} linear upsample         &  68.33  & 68.96   \\
        2-layer deconv \textit{w.} 2-layer upsample   &  67.51  & 67.65   \\
        \bottomrule
    \end{tabular}%
    }
    \end{center}
    \label{tab:ablation_head}
    \vspace{-15pt}	
\end{minipage}
\end{table*}

\vspace{-6pt}
\section{Experiments}
\label{4.0-experiments}
\vspace{-5pt}
\subsection{Experimental setups}
\vspace{-5pt}

\textbf{Datasets and evaluation metrics.} Our method is validated in the REC, RES, and PG tasks with five widely used datasets, namely three REC/RES datasets (RefCOCO/+/g \cite{yu2016modeling, mao2016generation}), as well as two PG datasets (ReferItGame \cite{kazemzadeh2014referitgame} and Flickr30k Entities \cite{plummer2015flickr30k}). In PG, the query pertains to a specific phrase, while in REC and RES, the query refers to a reference expression. The text of RefCOCO+/g exhibits greater length and complexity in comparison to that of RefCOCO. In REC/PG, we follow previous works \cite{deng2021transvg, resc} that employs Intersection-over-Union (IoU) as the evaluation metric, \ie, a prediction is deemed accurate only when its IoU exceeds or equals 0.5. We compute the prediction accuracy for each dataset as a performance indicator. While in RES, we follow previous works \cite{vg-law, kim2023risclip} that employs mean IoU (mIoU) and overall IoU (oIoU) for each dataset as the indicators. The detailed statistics information regarding these five datasets are provided in the \cref{sup_sec:intro_of_dataset}.

\noindent\textbf{Experimental details.} Since MRefM is proposed on the basis of the traditional MVLM, considering the pre-training cost of MVLM, we adopt BEiT-3 \cite{beit3} base and large model as our initial weights and then perform intermediate MRefM pre-training on the task-relevant dataset. Such intermediate pre-training is common in existing grounding works \cite{liu2023grounding, liu2023dqdetr, kamath2021mdetr}. As described in \cref{2.1-rec}, to verify the effectiveness of our MRefM approaches, \textbf{we conduct extensive experiments on three settings}: 
\textbf{(1) The basic single-dataset fine-tuning setting.} This setting does not require additional training data and aligns with existing supervised and self-supervised transfer approaches \cite{xiao2023clip, xiao2024hivg, wang2022cris, kim2023risclip}. In this setting, we perform supervised single-dataset intermediate MRefM pre-training before fine-tuning.  
\textbf{(2) The setting of fine-tuning with supervised dataset-mixed intermediate pre-training.} This setting aligns with existing grounded pre-trained approaches, such as Grounding-DINO \cite{liu2023grounding}, DQ-DETR \cite{liu2023dqdetr} \etc we perform an MRefM intermediate pre-training before fine-tuning.
\textbf{(3)} To verify the generality of MRefM, we perform \textbf{the setting of fine-tuning with unsupervised intermediate MRefM pre-training}. There are several ways to obtain the regions in the regional masking modeling works \cite{nguyen2023r-mae, geng2022multimodal, singh2022flava}, such as Felzenswalb-Huttenlocher (FH) algorithm \cite{felzenszwalb2004fh}, SAM \cite{kirillov2023segment} \etc  Thus, we adopt the unsupervised, fast, image-computable FH algorithm \cite{felzenszwalb2004fh} to generate regions following R-MAE \cite{nguyen2023r-mae}. We then select the referred one using a BEiT-3 model with performed image-text contrastive intermediate tuning. More details about the selection of the unsupervised regions, network architecture, training and inference, model hyperparameters \etc are provided in the \cref{sup_sec:imp_details}.

\vspace{-5pt}
\subsection{Comparison with state-of-the-art methods}
\vspace{-5pt}

\textbf{Referring expression comprehension.} As shown in \cref{tab:rec_sota1} and \cref{tab:rec_sota2}, we conducted experiments for the REC and PG tasks across \textbf{three settings}. \textbf{(1)} In the single-dataset fine-tuning setting, our base model surpasses the current SoTA method HiVG \cite{xiao2024hivg} by 2.07$\%$(testB), 6.15$\%$(testB), 4.73$\%$(test), 1.95$\%$(test), and 1.50$\%$(test) on the five datasets respectively, while also significantly outperforming the traditional uni-modal detector-based approach TransVG++ \cite{transvg++} by 4.37$\%$(testB), 7.98$\%$(testB), 7.22$\%$(test), 2.47$\%$(test), and 2.12$\%$(test), respectively. \textbf{(2)} In dataset-mixed pre-training setting, our base model outperforms HiVG \cite{xiao2024hivg} by 1.35$\%$, 2.79$\%$, and 2.63$\%$ on RefCOCO/+/g testB/testB/test splits, outperforms Grounding-DINO \cite{liu2023grounding} by 2.59$\%$, 4.76$\%$, and 2.38$\%$, exceeds OFA by 5.28$\%$, 5.18$\%$ ,and 5.01$\%$ , and even surpasses LION \cite{chen2024lion} - a GMLLM model that is 20-60 times larger than ours - by 3.76$\%$, 2.13$\%$ ,and 1.69$\%$. Note that among these works, UniTAB \cite{yang2022unitab}, OFA\cite{wang2022ofa}, LION \cite{chen2024lion} also utilize the MVLM on the pre-training stage. \textbf{(3)} Furthermore, we achieve competitive performance in the unsupervised setting, which shows the generality of MRefM paradigm. Additionally, our large-size model exhibits remarkable scalability with further substantial improvements in performance. More detailed results are provided in the \cref{sup_sec:experiment}.

\textbf{Referring expression segmentation.} As presented in \cref{tab:res_sota1} (mIoU metric), we conducted experiments for RES task under \textbf{three settings}. \textbf{(1)} In the single-dataset fine-tuning setting, our base model surpasses the SoTA self-supervised method RISCLIP \cite{kim2023risclip} by 2.65$\%$, 4.77$\%$, and 1.73$\%$ on RefCOCO/+/g testB/testB/test splits, respectively, while also significantly outperforming the traditional uni-modal detector-based approach VG-LAW \cite{vg-law} by 3.42$\%$, 7.31$\%$, and 4.57$\%$, respectively. \textbf{(2)} In the dataset-mixed pre-training setting, our base model achieves superior performance compared to the SoTA method RISCLIP \cite{kim2023risclip} with improvements of 4.53$\%$, 8.21$\%$, and 5.39$\%$. \textbf{(3)} In the unsupervised pre-training setting, we also achieve competitive performance. Additionally, our large-size model also exhibits remarkable scalability and demonstrates a substantial improvement in performance. For oIoU metric, the results are presented in \cref{sup_subsec:res_oiou} (\cref{tab:res_sota2}).

\begin{figure*}[t]
\vspace{-7pt}
 \centering
   \includegraphics[width=1.0\linewidth]{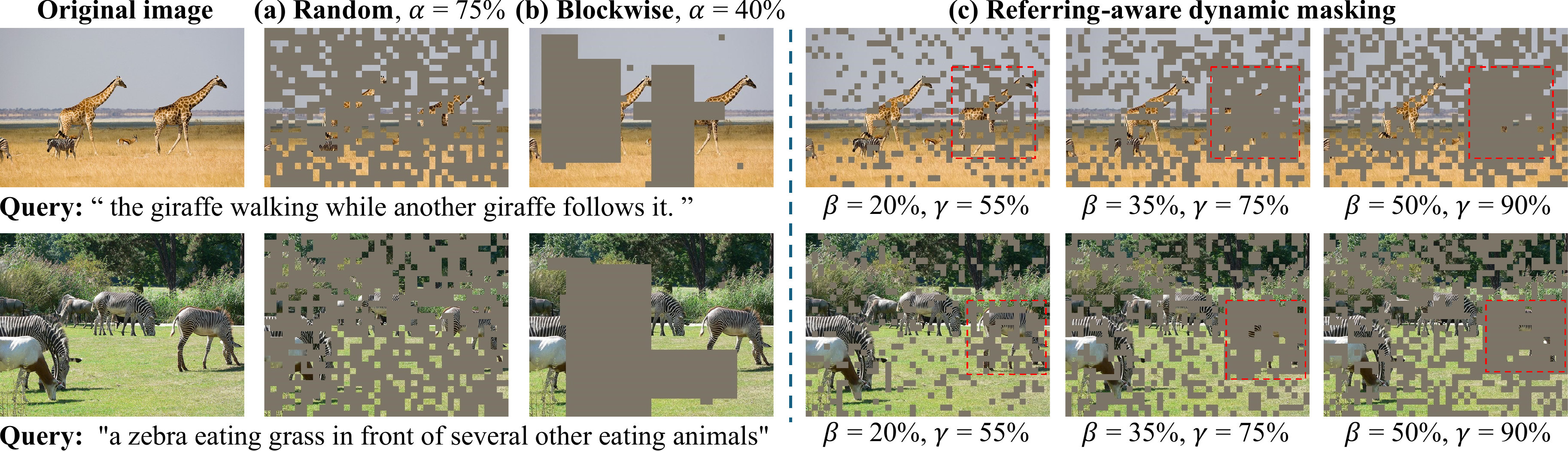}
   \vspace{-17pt}
   \caption{Illustrations of random masking (MAE) \cite{he2022masked}, block-wise masking (BEiT) \cite{bao2021beit}, and our referring-aware dynamic masking. $\alpha$ denotes the entire masking ratio, while $\beta$ and $\gamma$ denote the masking ratio beyond and within the referred region.}
    \vspace{-8pt}
   \label{fig:mask_sample}
\end{figure*}

\begin{table*}[t]
\centering
\begin{minipage}[t]{0.595\textwidth}
\centering
    \vspace{-5pt}
    \footnotesize
    \caption{Generality study of MRefM on RefCOCOg.}
    \centering
    \resizebox{1.0\columnwidth}{!}{%
    \begin{tabular}{l|c|cc|cc}
    \toprule
    \multirow{2}[1]{*}{Architecture} & \multirow{2}[1]{*}{Backbone} & \multicolumn{2}{c|}{Single-dataset} &  \multicolumn{2}{c}{Mixup pretrain}  \\
        &  & val  & test    &    val  & test                 \\
    \midrule
    TransVG         & DETR / BERT-B    & 68.67 & 67.73 & 75.73 & 75.86  \\  \rowcolor{cyan!03} 
    MRefM-TransVG     & DETR / BERT-B    & \textbf{71.51} & \textbf{70.84} & \textbf{78.71} & \textbf{78.69}   \\
    CLIP-VG         & CLIP-B / CLIP-B  & 73.18 & 72.54 & 78.67 & 78.54  \\  \rowcolor{cyan!03}
    MRefM-CLIP-VG     & CLIP-B / CLIP-B  & \textbf{74.22} & \textbf{74.50} & \textbf{80.48} & \textbf{80.83}  \\
    \bottomrule
    \end{tabular}%
    }
    \label{tab:general_mrm}%
    \vspace{-15pt}    
\end{minipage}
\hspace{1.5mm}
\begin{minipage}[t]{0.38\textwidth}
\centering
    \footnotesize
    \vspace{-4pt}
    \caption{Generality of the task heads.}
    \vspace{-13pt}
    \begin{center}
    \resizebox{1.0\columnwidth}{!}{%
    \begin{tabular}{l|c|c}
        \toprule
        \multirow{2}[1]{*}{Architecture  (Fine-tuning setting)}   &   \multicolumn{2}{c}{RefCOCOg}  \\
                                                &   val  & test  \\
        \midrule
        TransVG++ (Reproduced by us)           &  75.04  & 75.55  \\
        \rowcolor{cyan!03}
        TransVG++ \textit{w.} our REC head     &  \textbf{76.65}  & \textbf{77.09}   \\
        \midrule                                                           %
        LAVT \cite{lavt}                       &  63.34  & 63.62   \\
        \rowcolor{cyan!03}
        LAVT \textit{w.} our RES head          &  \textbf{64.84}  & \textbf{65.35}   \\
        \bottomrule
    \end{tabular}%
    }
    \end{center}
    \label{tab:general_head}
    \vspace{-15pt}	
\end{minipage}
\end{table*}

\vspace{-3pt}
\subsection{Ablation study}
\vspace{-3pt}

\textbf{The Mask Referring Modeling.} In \cref{tab:ablation_mrm}, we conducted ablation studies on MRefM, which included Referring MIM (`Ref MIM'), Referring MLM (`Ref MLM'), and referring-aware dynamic image masking (`referring-aware'). The `vanilla' denotes the vanilla MVLM described in \cref{3.1-preliminary}. As shown in \cref{tab:ablation_mrm}, referring MIM, referring MLM, and dynamic masking strategy resulted in improvements of 3.70$\%$, 2.16$\%$, and 1.05$\%$ on the RefCOCOg-test dataset, and with an overall improvement of 6.21$\%$, demonstrates the effectiveness of our methods. More results are provided in the \cref{sup_subsec:Complete_ablation_study_of_MRefM}.

\textbf{The referring-aware dynamic masking strategy.} 
\cref{fig:mask_sample} presents a schematic of the three masking strategies. In our experiments, as illustrate in \cref{fig:mask_sample}-(c), $\beta$ and $\gamma$ demonstrate optimal performance at values of 0.35 and 0.75, respectively. More detailed results are provided in the \cref{sup_subsec:Complete_ablation_study_of_dynamic_masking}.

\textbf{The referring-based task heads.} We conducted ablation studies on the design of two referring-based task heads. \cref{tab:ablation_head} reveals that our modeling method effectively captures referring information at the backbone stage, benefiting from the one-tower structure. This approach is significantly more efficient than the traditional fusion encoder and special token-based method. Additionally, our proposed box mask loss also contributes to a performance gain of 1.50$\%$(test).

\vspace{-3pt}
\subsection{Generality study}
\vspace{-3pt}

\textbf{The generality of MRefM.} Firstly, we perform an unsupervised MRefM pre-training in \cref{tab:rec_sota2} and \cref{tab:res_sota1}, both of which achieve competitive performance. Secondly, we replace the backbone and apply MRefM on DETR and CLIP by using TransVG \cite{deng2021transvg} and CLIP-VG \cite{xiao2023clip} under the two settings. Since the two frameworks do not interact at backbone stage, we build MRefM on the fusion encoder. In \cref{tab:general_mrm}, MRefM can effectively learn referring representation, resulting in an overall performance gain of about 2.0$\%$. All these findings demonstrate the validity and generality of the MRefM paradigm. 

\textbf{The generality of referring-based task heads.} Since both TransVG++ \cite{transvg++} and LAVT \cite{lavt} have modality interactions at backbone stage, we attempted to apply our task heads to both frameworks. TransVG++ is reproduced by us since its code is not available. \cref{tab:general_head} shows that our proposed task heads achieve a 1.5+$\%$ improvement in both REC and RES, offering a new avenue for future research.

\vspace{-5pt}
\section{Conclusion}
\label{5.0-conclusion}
\vspace{-5pt}

In this paper, we propose a novel, highly concise, and feature space unified one-tower referring framework. Additionally, we pioneer the exploration of mask modeling in referring tasks by introducing MRefM paradigm to capture the referential relationships between vision and text. We demonstrate the effectiveness and generality of MRefM across three settings on REC, PG, and RES tasks, consistently achieving groundbreaking results. Furthermore, leveraging unsupervised methods enables potential large-scale pre-training of MRefM in the future, presenting a new direction for referring tasks.

\section*{Acknowledgement}

This work was supported in part by the National Natural Science Foundation of China under Grants 62036012, U23A20387, 62322212, 62072455, in part by Pengcheng Laboratory Research Project under Grant PCL2023A08, in part by National Science and Technology Major Project under Grant 2021ZD0112200, in part by Alibaba Innovative Research Program, and also in part by CAS Project for Young Scientists in Basic Research (YSBR-116).


\medskip

{\small
\bibliographystyle{plainnat}
\bibliography{ref}
}

\newpage
\appendix



\begin{center}
{\fontsize{14pt}{1pt}\selectfont
\textbf{Appendix}
}
\end{center}


$\ \ \ $We provide an overview of the Appendix below:

\begin{itemize}
    \item \textbf{\cref{sup_sec:explanation_of_task_definition}: Explanation of the task definition}

    \item \textbf{\cref{sup_sec:intro_of_dataset}: Introduction of the datasets} 
        \begin{itemize}
            \item \cref{sup_subsec:intro_of_dataset} The five referring datasets.
            \item \cref{sup_subsec:intro_of_dataset_abbre}  Explanation of the dataset abbreviations.
            \item \cref{sup_subsec:compare_of_dataset}  Comparison of datasets used in the pre-trained models.
        \end{itemize}    
    
    \item \textbf{\cref{sup_sec:imp_details}: Implementation details}
           
    \item \textbf{\cref{sup_sec:technical}: Technical remarks}
        \begin{itemize}
            \item \cref{sup_subsec:effectiveness_mechanism_for_visual_relation_score} Further explanation for the effectiveness mechanism of the visual target-relation score.
            \item \cref{sup_subsec:selection_of_unsupervised_region}  The selection of the unsupervised regions.
            \item \cref{sup_subsec:referring-aware_text_masking} Referring-aware text masking.
            \item \cref{sup_subsec:difference_of_task_heads} The difference of the task heads between ours with other frameworks.
        \end{itemize}
        
    \item \textbf{\cref{sup_sec:experiment}: Extra experimental results}
        \begin{itemize}
            \item \cref{sup_subsec:ReferIt_and_Flickr30k} The results on phrase grounding task under mixup pre-training setting.
            \item \cref{sup_subsec:computational_costs_analysis} Computational costs analysis compared with SoTA methods.
            \item \cref{sup_subsec:Complete_ablation_study_of_MRefM} Complete ablation study of MRefM.
            \item \cref{sup_subsec:Complete_ablation_study_of_dynamic_masking} Ablation study of the mask ratio in referring-aware dynamic masking.
        \end{itemize}    
    
    \item \textbf{\cref{sup_sec:visualize}: Visualization of the results}
    
    \item \textbf{\cref{sup_sec:further_discuss}: Further discussions}
        \begin{itemize}
            \item \cref{sup_subsec:Limitations} Limitations.
            \item \cref{sup_subsec:Broader_impacts} Broader impacts.
        \end{itemize}   
    
\end{itemize}

\section{Explanation of the task definition}
\label{sup_sec:explanation_of_task_definition}

As explained in \cref{1.0-introduction} of the main text, Visual Grounding (VG) aims to grounding a region referred by a query text in a specific image. The generalized visual grounding includes Referring Expression Comprehension (REC), Phrase Grounding (PG), and Referring Expression Segmentation (RES) tasks. However, in recent years, REC and RES have often been studied separately. Therefore, in numerous works \cite{deng2021transvg, transvg++, vg-law, xiao2024hivg, xiao2023clip, qrnet}, visual grounding specifically refers to REC and PG tasks, which involve grounding a rectangular region. In this paper, we follow the mainstream and have not clearly separated the “grounding” from “generalized visual grounding” and “REC and PG tasks”. When expressing the experimental task, “grounding” usually refers to REC or PG tasks, so as to discuss it parallelly with the RES task.

\section{Introduction of the datasets}
\label{sup_sec:intro_of_dataset}

\subsection{The five referring datasets}
\label{sup_subsec:intro_of_dataset}

We present the detailed descriptions of the five referring datasets used in our experimental study on the REC, PG, and RES tasks. \cref{tab:dataset_st} presents the detailed statistics.

\noindent\textbf{RefCOCO/RefCOCO+/RefCOCOg.} These three datasets belong to the Referring Expression Comprehension (REC) and Referring Expression Segmentation (RES) tasks, and the images of these three datasets derived from MSCOCO~\cite{mscoco}. Expressions in RefCOCO~\cite{yu2016modeling} and RefCOCO+~\cite{yu2016modeling} are collected by the two-player game proposed in ReferitGame~\cite{kazemzadeh2014referitgame}. There are two test splits called ``testA'' and ``testB''. Images in ``testA'' only contain multiple people annotation. In contrast, images in ``testB'' contain all other objects. Expressions in RefCOCOg~\cite{mao2016generation} are collected on Amazon Mechanical Turk in a non-interactive setting. Thus, the expressions in RefCOCOg are longer and more complex. RefCOCOg has ``google'' and ``umd'' splits. \textbf{The ``google'' split} does not have a public test set, and \textbf{exists an overlap between the training and validation image sets.} The ``umd'' split does not have this overlap. 
Therefore, \textbf{to prevent data leakage of the test set} and following previous studies \cite{vg-law, yu2018mattnet}, \textbf{we exclude the ``google'' split in the fine-tuning settting and dataset-mixed pre-training setting}. Thus, we trained and tested the RefCOCOg dataset only on the ``umd'' split.

\noindent\textbf{ReferItGame.} ReferItGame~\cite{kazemzadeh2014referitgame} (short as ReferIt) belongs to the Phrase Grounding (PG) task, which contains images from SAIAPR12~\cite{escalante2010segmented} and collects expressions through a two-player game. In this game, the first player is shown an image with an object annotation and is asked to write a natural language expression referring to the object. The second player is then shown the same image along with the written expression and is asked to click on the corresponding area of the object. If the clicking is correct, both players receive points and swap roles. If not, a new image will be presented.

\noindent\textbf{Flickr30k Entities.} Flickr30k Entities (short as Flickr30k) ~\cite{plummer2015flickr30k} belongs to the phrase grounding task, which contains images in Flickr30k dataset. The query sentences are short noun phrases in the captions of the image. The queries are simpler and easier to understand compared to RefCOCO/+/g. Therefore, the ambiguity of the expression is heightened simultaneously, resulting in a relative increase in noise.

\begin{table*}[t]
  \vspace{-5pt}
  \centering
  \caption{The detailed statistics of RefCOCO \cite{yu2016modeling}, RefCOCO+ \cite{yu2016modeling}, RefCOCOg \cite{mao2016generation}, ReferItGame \cite{kazemzadeh2014referitgame} and Flickr30K Entities \cite{plummer2015flickr30k} datasets. We represent test split and testA split in the same column.}
  \resizebox{0.8\columnwidth}{!}{%
  \begin{tabular}{c|ccccccc}
  \toprule
  \multirow{2}{*}{Dataset}  & \multirow{2}{*}{Images} & \multirow{2}{*}{Instances} & total   & train   & val   & test(A)   & testB   \\
                            &                         &                            & queries & queries & queries & queries & queries \\ \midrule
  RefCOCO \cite{yu2016modeling}                   & 19,994 & 50,000    & 142,210  & 120,624  & 10,834  & 5,657 & 5,095 \\
  RefCOCO+\cite{yu2016modeling}                  & 19,992 & 49,856    & 141,564  & 120,191  & 10,768  & 5,726 & 4,889 \\
  RefCOCOg \cite{mao2016generation}             & 25,799 & 49,822    & 95,010   & 80,512  & 4,896  & 9,602 & -- \\
  ReferItGame\cite{kazemzadeh2014referitgame}    & 20,000 & 19,987    & 120,072  & 54,127  & 5,842  & 60,103 & -- \\
  Flickr30k \cite{plummer2015flickr30k}  & 31,783 & 427,000   & 456,107  & 427,193  & 14,433  & 14,481 & -- \\
  \bottomrule
  \end{tabular}
  }
  \label{tab:dataset_st}
\end{table*}

\begin{table*}[t]
  \centering
  \caption{Comparison of datasets used in the pre-trained models of the comparable methods.}
  \vspace{+4pt}
  \resizebox{1.01\columnwidth}{!}{%
  \begin{tabular}{c|cccc}
  \toprule
  Pretrained model  & Uni-modal image / Data size & Uni-modal text / Data size   & Image-text pairs /  Data size   & Total   \\
  \midrule
  CLIP \cite{radford2021learning}    & --     & --   & LAION-400M \cite{schuhmann2021laion} / 400M & 400M \\
  UniTAB \cite{wang2023one}   &  \multicolumn{3}{c}{image-text pairs: CC3M\cite{changpinyo2021cc12m}, \etc; image-text-box pairs: COCO, VG, RefC, O365, SUB, \etc from multiple downstream task}    & >20M \\
  
  OFA \cite{wang2022ofa}   &  ImageNet-21K,\etc /  40M    &  filtered BookCorpus\cite{zhu2015aligning}\etc  / 140GB  & CC12M\cite{changpinyo2021cc12m}, SBU\cite{ordonez2011sbu}, COCO\cite{mscoco}, VG\cite{krishna2017visualgenome},\etc  / 21M & 40M+140GB+21M \\
  EVA-G \cite{fang2023eva}    &  --      &  --   & Merged-2B(LAION-2B\cite{schuhmann2021laion}+COYO-700M)  / >2B & >2000M \\
  FlanT5 \cite{fang2023eva}    &  --      &  filtered crawl data  / >750GB  & --   & >750GB \\
  ONE-PEACE \cite{wang2023one}   &  \multicolumn{3}{c}{image-text pairs: LAION-2B \cite{schuhmann2021laion}; audio-text pairs: 2.4M + 8000 hours}    & >2000M \\
  \rowcolor{pink!19}
  BEiT-3 \cite{beit3}    &  ImageNet-21K\cite{imagenet} /   14M    &  BookCorpus\cite{zhu2015aligning},\etc  / 160GB  & CC12M\cite{changpinyo2021cc12m}, SBU\cite{ordonez2011sbu}, COCO\cite{mscoco}, VG\cite{krishna2017visualgenome},\etc  / 21M & 14M+160GB+21M \\
  \bottomrule
  \end{tabular}
  }
  \label{tab:model_data_st}
  \vspace{-5pt}
\end{table*}

\subsection{Explanation of the dataset abbreviations}
\label{sup_subsec:intro_of_dataset_abbre}

In \cref{tab:rec_sota2} of the main text, we provide abbreviations for the datasets used in intermediate pre-training. Specifically, `GoldG' (proposed in MDETR \cite{kamath2021mdetr}) is a mixed region-level fine-grained dataset created by combining three datasets - Flickr30k \cite{plummer2015flickr30k}, MS COCO \cite{mscoco}, and Visual Genome \cite{krishna2017visualgenome} - along with annotated text data for detection, REC and QGA tasks. It has a size of approximately $6.2$M. `O365' refers to the Object365 \cite{zhou2019objects} dataset, `SBU' stands for SBU caption \cite{ordonez2011sbu}, `VG' represents the Visual Genome \cite{krishna2017visualgenome} dataset, and `OI' stands for OpenImage \cite{kuznetsova2020openimage} dataset.

\subsection{Comparison of datasets used in the pre-trained models}
\label{sup_subsec:compare_of_dataset}

As presented in \cref{tab:model_data_st}, we conducted an analysis of the datasets employed by the backbone models compared in \cref{tab:rec_sota2} within the dataset-mixed pre-training setting. From the \cref{tab:model_data_st}, it is evident that BEiT-3 and OFA utilize comparable datasets for pre-training. Conversely, other compared works in \cref{tab:rec_sota2}, such as Shikra \cite{chen2023shikra}, Ferret \cite{you2023ferret}, LION \cite{chen2024lion}, and other models, such as ONE-PEACE \cite{wang2023one} (a tri-modality foundation model), employ significantly larger amounts of data than BEiT-3. Consequently, our method does not possess any advantage concerning the volume of data used in pre-training.

\begin{table*}[t]
\footnotesize
\caption{Network structure of our proposed OneRef framework.}
\vspace{-4pt}
\begin{center}
\resizebox{0.78\columnwidth}{!}{%
\begin{tabular}{l|c|c|ccc|c}
    \toprule
\multirow{2}{*}{Model} & \multirow{2}{*}{Backbone} &  \multirow{1}{*}{Input} & \multicolumn{3}{c|}{One-tower Transformer}  & All parameters  \\
 &   & resolution & layers & dimension & heads  &  (include all MoE heads) \\
    \midrule
OneRef-B           & BEiT-B/16 &  384 & 12 & 768  & 12 & 234M (REC), 267M (RES)  \\
OneRef-L           & BEiT-L/16 &  384 & 24 & 1024 & 16 & 639M (REC), 679M (RES) \\
    \bottomrule
\end{tabular}%
}
\end{center}
\label{tab:params}
\vspace{-7pt}	
\end{table*}

\section{Implementation details}
\label{sup_sec:imp_details}

\noindent\textbf{Network Architecture.} The detailed network structure of our framework is shown in \cref{tab:params}. We employ BEiT-B/16 and BEiT-L/16 as the backbone for our OneRef base and large version, respectively. In the structure of OneRef-B, the one-tower encoder are 12-layer Transformers with the hidden embedding dimension of 768. In the structure of OneRef-L, the one-tower encoder is 24-layer Transformers with the hidden embedding dimension of 1024. The one-tower encoder encodes both textual and visual modalities. Due to the utilization of a 3-layer deconvolution, RES exhibits a slightly higher number of model parameters compared to the REC task.

\noindent\textbf{Training Details.}  The batch size for pre-training the base model and large model are (32, 8), while they are (32, 8) and (16, 6) for transferring to the REC and RES tasks, respectively. Our model is optimized end-to-end by using the AdamW optimizer and a cosine learning scheduler with an initial learning rate of $0.5\times10^{-4}$ for 110 epochs during the pre-training stage. During REC/RES transfer stage, the learning rates is $0.3\times10^{-4}$ with 20 epochs. The framework and experiments in our study were conducted using PyTorch. For MRefM pre-training, the base model took 15 hours on 32 NVIDIA A100 GPUs, while the large model took 50 hours on the same number of GPUs. As for REC/RES transfer fine-tuning training, it took an average of 3 hours for the base model and 8 hours for the large model to process one dataset on 8 A100 GPUs.

\noindent\textbf{Inference Details.} Unlike previous methods, such as TransVG++ \cite{transvg++}, QRNet \cite{qrnet}, \etc, which heavily rely on high-resolution images like 640$\times$640, we adopt smaller resolution of 384$\times$384. To ensure compatibility, we employ a long edge alignment and short edge pad filling scheme to the image. We include \texttt{[SEP]} and \texttt{[EOS]} token at the beginning and the end of the input text, and align it to a fixed length of 64 by padding empty tokens.

\noindent\textbf{Model Hyperparameters.} We summarize and report the hyperparameter settings of the OneRef framework in \cref{tab:hyperparam}.

\begin{table}[t]
\footnotesize
\caption{Hyperparameters of our framework during training. \textit{lr.} denotes the learning rate.} 
\vspace{-10pt}
\begin{center}
\resizebox{0.57\columnwidth}{!}{%
\begin{tabular}{l|c|c}
    \toprule
    \multirow{2}{*}{Item} &  \multicolumn{2}{c}{Value}  \\
                          & base model & large model  \\
    \midrule
    optimizer                                 &    \multicolumn{2}{c}{AdamW}         \\
    Epoch for MRefM pre-training              &    \multicolumn{2}{c}{110}         \\
    \textit{lr.} for MRefM pre-training       &    \multicolumn{2}{c}{0.5$\times$$10^{-4}$}  \\
    weight decay                              &    \multicolumn{2}{c}{0.5$\times$$10^{-5}$}   \\
    patch size                                &    \multicolumn{2}{c}{16$\times$16}   \\
    Initial value of aspect ratio $a$ in MIM  &    \multicolumn{2}{c}{0.3}   \\
    mask ratio $\beta$ in Referring MIM       &     \multicolumn{2}{c}{0.75} \\
    mask ratio $\gamma$ in Referring MIM      &     \multicolumn{2}{c}{0.35} \\
    mask ratio $\delta$ in Referring MLM      &     \multicolumn{2}{c}{0.40} \\
    $\lambda_{reg}$ and $\lambda_{img}$ in Referring MLM     &     \multicolumn{2}{c}{1, 1} \\
    batch size for MRefM pre-training         &     32  &  8 \\
    \midrule
    Epoch for REC/RES transfer                &    \multicolumn{2}{c}{20}         \\
    \textit{lr.} for REC/RES transfer         &    \multicolumn{2}{c}{0.3$\times$$10^{-4}$}  \\
    $\lambda_{l_1},\  \lambda_{giou}$ in REC  &   \multicolumn{2}{c}{2, 2}     \\
    $\lambda_{f},\ \lambda_{d}$ in REC        &  \multicolumn{2}{c}{20, 2}     \\
    $\lambda_{seg\_f},\ \lambda_{seg\_d}$ in RES  &  \multicolumn{2}{c}{20, 2}  \\
    batch size for REC/RES transfer          &     32 / 16  &  8 / 6 \\
    \bottomrule
\end{tabular}%
}
\end{center}
\label{tab:hyperparam}
\vspace{-8pt}	
\end{table}

\section{Technical remarks}
\label{sup_sec:technical}

\begin{figure*}[h!]
 \centering
   \includegraphics[width=0.92\linewidth]{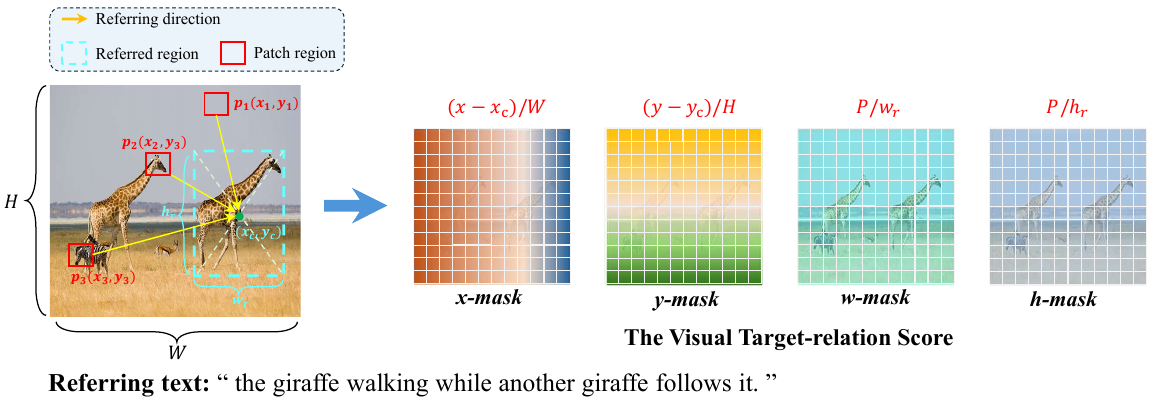}
   \vspace{-7pt}
   \caption{The reconstruction of the visual target-relation score $\bm{\vs}^{vt}  \in \mathbb{R}^{N_v \times 4}$. $(x,y)$ represents the coordinates of a general image patch, and $P$ is the patch size. By slicing the predicted score, four masks can be derived. The score represents the spatial distance and the relative size between the current patch region and the referred region.}
   \label{fig:score_explanation}
    \vspace{-12pt}
\end{figure*}

\subsection{Further explanation for the effectiveness mechanism of the visual target-relation score}
\label{sup_subsec:effectiveness_mechanism_for_visual_relation_score}

\textbf{(1) The purpose of designing the Referring MIM algorithm.} In the existing MIM paradigm, reconstruction is limited to solely relying on the visual features within the image. To enhance content reconstruction by leveraging cross-modal information as much as possible, our Referring MIM approach incorporates visual target-relation scores alongside visual modality content during reconstruction. This modeling approach presents increased difficulty as it necessitates reliance on textual information for reconstructing the two visual branch. Consequently, our model achieves a more comprehensive understanding of both visual and textual information. In this way, the model not only can perceive the information of the image modality itself but also have a more accurate understanding of the location and correlation of the key object features in different regions.

\textbf{(2) How and why the visual target-relation score (\ie, the \textit{x-, y-, w-,} and \textit{h-masks}) works.} We provide a clearer illustration in \cref{fig:score_explanation} for further explanation. As mentioned in \cref{subsec:ref_mim}, this score represents the spatial distance between the current patch region and the referred region, it enables implicit deployment of grounding capability within each token of the model. When reconstructing the visual features and target-relation score of each local patch, the model actually needs to have an global and comprehensive understanding of the text modality information and the visual information. On this basis, the model needs to rely on the reconstructed visual features of the local patch to implicitly predict the specific location and size of the referred object, and then accurately predict the visual target-relation score. Finally, Referring MIM can enhance the model's global and multimodal understanding of textual and visual information, and then learn more general visual representations, which can have better generalization ability when deployed to downstream referring tasks.

The proposed Referring MIM is our own design, which is mainly used to improve the defects existing in MAE \cite{he2022masked}/BEiT \cite{bao2021beit}. We can find the rationale of our method in some classic computer vision works, such as the YOLO series works \cite{redmon2016yolo}, which predicts the location, size, confidence, and category of the object box corresponding to each grid cell based on the global understanding of the image. YOLO \etc \cite{redmon2016yolo} also confirmed that the object detection model obtained in this way has stronger generalization ability when transfer to detection tasks that differ greatly from the training data compared with other detectors.

\subsection{The selection of the unsupervised regions}
\label{sup_subsec:selection_of_unsupervised_region}

The process of selecting unsupervised regions bears resemblance to weakly-supervised visual grounding. Drawing inspiration from ALBEF's method \cite{li2021albef} for weakly-supervised grounding, we employ a BEiT-3 model with performed image-text contrastive tuning to encode both the image and text, thereby obtaining a cross-modal text-to-image attention map for selection. Subsequently, leveraging the cross-modal attention and modular parsing of textual sentences provided by MAttNet \cite{yu2018mattnet} enables us to derive scores for each proposal. Finally, we select the region with the highest score as our objective in Referring MRefM.

\subsection{Referring-aware text masking}
\label{sup_subsec:referring-aware_text_masking}

In referring MLM, we utilize a referring-aware text masking strategy. Specifically, we preferentially mask out the referential subject of the expression text on the basis of a random mask, and the subject is obtained by the NLP parsing tool (\eg, spaCy). Since this small technical point does not observe a significant performance gain as the referring-aware dynamic image masking strategy, we do not provide additional ablation experiments.

\subsection{The difference of the task heads between ours with other frameworks}
\label{sup_subsec:difference_of_task_heads}

Recently, several multi-task visual grounding studies \cite{vg-law, li2021referring} have incorporated both grounding and segmentation task heads into their frameworks. Most relevance to our work is VG-LAW \cite{vg-law}, which simplifies the implementation of grounding and segmentation heads by eliminating Transformer-based fusion encoders through visual adaptive weights generation. In contrast, for REC headers, we propose a box mask constraint based on cross-modal cosine similarity that significantly enhances the accuracy of such grounding approach. For the RES head, instead of employing adaptive weights generation, we directly obtain segmentation masks using cosine similarity for the visual tokens upsampled by a 3-layer deconvolution.

\section{Extra experimental results}
\label{sup_sec:experiment}

\begin{table*}[t]
\vspace{-5pt}
    \footnotesize
    \caption{Comparison with \textbf{latest} SoTA methods for PG task with dataset-mixed intermediate pre-training setting. `RefC' represents the mixup of RefCOCO/+/g training data. $\dagger$ indicates RefC has been used during pre-training.}
    \centering
    \resizebox{0.88\columnwidth}{!}{%
    \begin{tabular}{c|c|c|c|c|c|c}
    \toprule
    \multirow{2}[1]{*}{Methods} & \multirow{2}[1]{*}{Venue} & Visual/Language & Intermediate  & Data & \multicolumn{1}{c|}{ReferIt}  & \multicolumn{1}{c}{Flickr}   \\
                                &                           & Backbone        & pretrain data & size  & test    & test        \\
    \midrule       
    \multicolumn{7}{l}{\textbf{Dataset-mixed intermediate pre-training setting}}       \\ 
     MDETR $^\dagger$ \cite{kamath2021mdetr}     & ICCV'21   & RN101/RoBERT-B        & GoldG,RefC             & 6.5M  & --    & 83.80 \\  
     YORO$^\dagger$ \cite{ho2023yoro}  & ECCV'22 & ViLT \cite{kim2021vilt} / BERT-B  & GoldG,RefC             & 6.5M  & 71.90 & --    \\
     UniTAB $^\dagger$ \cite{yang2022unitab}     & ECCV'22   & RN101/RoBERT-B        & VG,COCO,\etc           & >20M  & -- &  79.38   \\
     HiVG-B$^\dagger$ \cite{xiao2024hivg}        & ACMMM'24  & CLIP-B / CLIP-B       & RefC,ReferIt,Flickr    & 0.8M  & 77.75 & 82.08 \\
     HiVG-L$^\dagger$ \cite{xiao2024hivg}        & ACMMM'24  & CLIP-L / CLIP-L       & RefC,ReferIt,Flickr    & 0.8M  & 78.16 & 82.63  \\
        \midrule
    \rowcolor{cyan!03}
    \textbf{OneRef-B$^\dagger$ (0.2B)}        &   NeurIPS'24     & BEiT3-B / BEiT3-B      & RefC,ReferIt   & 0.5M     &   \textcolor{red}{79.66}    &   \textcolor{red}{84.01}        \\
    \rowcolor{cyan!03}
    \textbf{OneRef-L$^\dagger$ (0.6B)} &  NeurIPS'24   & BEiT3-L / BEiT3-L &  RefC,ReferIt  & 0.5M    &  \textbf{83.22} & \textbf{85.13}   \\
    \bottomrule
    \end{tabular}%
    }
    \label{tab:rec_sota_pg}%
    \vspace{-5pt}    
  \end{table*}%

\begin{table*}[t!]
\vspace{-3pt}
    \footnotesize
    \caption{Comparison with \textbf{latest} SoTA methods (\textbf{oIoU} metric) on the three datasets for \textbf{RES} task with both single-dataset fine-tuning setting and dataset-mixed intermediate pre-training setting. $\dagger$ indicates RefC has been used during pre-training.}
    \centering
    \resizebox{1.0\columnwidth}{!}{%
    \begin{tabular}{c|c|c|c|ccc|ccc|cc}
    \toprule
    \multirow{2}[1]{*}{Methods} & \multirow{2}[1]{*}{Venue} & Visual/Language & Intermediate & \multicolumn{3}{c|}{RefCOCO} & \multicolumn{3}{c|}{RefCOCO+} & \multicolumn{2}{c}{RefCOCOg}  \\
                                &                           & Backbone        & pretrain data &   val  & testA & testB       &   val  & testA & testB        &   val  & test                  \\
    \midrule
    \multicolumn{12}{l}{\textbf{Single-dataset fine-tuning setting \textit{w.} uni-modal pre-trained close-set segmentation model: (traditional setting)}}        \\
    LAVT \cite{lavt}                &   CVPR'22  & Swin-B / BERT-B   &  --    & 72.73  & 75.82 & 68.79 &  62.14 & 68.38 & 55.10 & 61.24 & 62.09  \\
    \midrule  
    \multicolumn{12}{l}{\textbf{Single-dataset fine-tuning setting \textit{w.} vision-language self-supervised pre-trained model: }}       \\ 
    RISCLIP-B \cite{kim2023risclip} &  NAACL'24  & CLIP-B / CLIP-B &  --    & 73.57  & 76.46 & 69.76 &  65.53 & 70.61 & 55.19 & 64.10 & 65.09  \\
    RISCLIP-L \cite{kim2023risclip} &  NAACL'24  & CLIP-L / CLIP-L &  --    & 76.92  & 80.99 & 73.04 &  71.24 & 76.99 & 61.56 & 67.96 & 68.71  \\
    \midrule
    \rowcolor{cyan!03}
    \textbf{OneRef-B (ours)}        &   NeurIPS'24      & BEiT3-B / BEiT3-B & --      &   \textcolor{red}{77.55}    &   \textcolor{red}{80.96}    &   \textcolor{red}{73.53}    &   \textcolor{red}{70.82}    &   \textcolor{red}{74.53}    &   \textcolor{red}{64.06}    &   \textcolor{red}{70.68}    &   \textcolor{red}{70.61}      \\
    \rowcolor{cyan!03}
    \textbf{OneRef-L (ours)} &   NeurIPS'24   & BEiT3-L / BEiT3-L &  --    &  \textbf{80.48} & \textbf{82.78} & \textbf{78.27}  & \textbf{74.25}  & \textbf{78.41}  & \textbf{69.85}  &  \textbf{74.91} &  \textbf{77.36}  \\
    \midrule
    \midrule
    \multicolumn{12}{l}{\textbf{Dataset-mixed intermediate pre-training setting:}}       \\ 
    HIPIE$^\dagger$ \cite{wang2024hierarchical}    &  NeurIPS'23 & RN50,CLIP / BERT-B & RefC,O365,PACO    & 78.30 & --    & --    & 66.20 & --    & --    & 69.80 & --  \\
    UNINEXT$^\dagger$ \cite{yan2023universal}   &  CVPR'23  & RN50 / BERT-B  & RefC,O365,SOT  & 77.90 & 79.68 & 75.77 & 66.20 & 71.22 & 59.01 & 70.04 & 70.52  \\
    PolyFormer-B$^\dagger$\cite{liu2023polyformer} &  CVPR'23  & Swin-B / BERT-B &  RefC    & 74.82 & 76.64 & 71.06 & 67.64 & 72.89 & 59.33 & 67.76 & 69.05  \\
    PolyFormer-L$^\dagger$\cite{liu2023polyformer} &  CVPR'23  & Swin-L / BERT-B &  RefC    & 75.96 & 78.29 & 73.25 & 69.33 & 74.56 & 61.87 & 69.20 & 70.19  \\
        \midrule  
    \rowcolor{cyan!03}
    \textbf{OneRef-B$^\dagger$ (ours)}        &   NeurIPS'24    & BEiT3-B / BEiT3-B & RefC      &   \textcolor{red}{81.06}    &   \textcolor{red}{83.05}    &   \textcolor{red}{77.80}    &   \textcolor{red}{72.24}    &   \textcolor{red}{77.32}    &   \textcolor{red}{67.08}    &   \textcolor{red}{75.14}    &   \textcolor{red}{77.21}      \\
    \rowcolor{cyan!03}
    \textbf{OneRef-L$^\dagger$ (ours)} &  NeurIPS'24   & BEiT3-L / BEiT3-L &  RefC    &  \textbf{81.95} & \textbf{83.62} & \textbf{79.57}  & \textbf{74.55}  & \textbf{79.54}  & \textbf{70.65}  &  \textbf{76.53} &  \textbf{79.18}  \\
    \bottomrule
    \end{tabular}%
    }
    \label{tab:res_sota2}%
    \vspace{-5pt}    
\end{table*}%

\subsection{The results on ReferIt and Flickr30k dataset under mixup pre-training setting}
\label{sup_subsec:ReferIt_and_Flickr30k}

The results of our framework on the PG task (\ie, ReferIt \cite{kazemzadeh2014referitgame} and Flickr30k \cite{plummer2015flickr30k} datasets) under the mixup pre-training setting are presented in \cref{tab:rec_sota_pg}. It is worth noting that a majority of studies conducted under this setting have not provided these results, thus only several works are included in the table. As shown in \cref{tab:rec_sota_pg}, our base model outperforms HiVG by 1.91$\%$ and 1.93$\%$ on the two datasets, and also achieves SoTA performance.

\subsection{The results for RES task under oIoU metric}
\label{sup_subsec:res_oiou}

The results for RES task under oIoU metric are presented in \cref{tab:res_sota2}. oIoU is calculated as the ratio between the total intersection area and the total union area of all test samples, each of which is consists of both a text query and an image. This metric particularly favors larger objects. As indicated in \cref{tab:res_sota2}, (1) in the single-dataset fine-tuning setting, our base model outperforms RISCLIP \cite{kim2023risclip} by 3.77$\%$, 8.87$\%$, and 5.52$\%$ on RefCOCO/+/g testB/testB/test split, respectively. (2) Similarly, in the dataset-mixed intermediate pre-training setting, our base model surpasses UNINEXT \cite{yan2023universal} by 2.03$\%$, 8.07$\%$, and 6.69$\%$ on RefCOCO/+/g testB/testB/test split, respectively. Furthermore, our large model exhibits remarkable scalability with additional performance enhancements.

\begin{table*}[t!]\footnotesize
\caption{Comparison of the computational cost in REC task. The results are obtained on RefCOCO dataset. The testing environment is 1 NVIDIA A100 GPU. $\dag$ indicates that the model's code is not publicly available, and the replicated estimation results are shown. The backbone parameters of our UniRef model only include the actual calculated parameters, specifically those of the V-L expert head in MoE, while excluding the parameters of unused visual and language expert heads and their uni-modal branches. We highlight the best result in \textbf{bold}. (\textit{FPS: images / (GPU $\cdot$ second)})}
\vspace{-3pt}
\begin{center}
\resizebox{0.92\columnwidth}{!}{%
    \begin{tabular}{c|c|c|c|c|c|c|c|c}
    \toprule
    \multirow{2}{*}{Model} & Backbone & Fusion+head  & Total  & FLOPs            & Fine-tune        &  Test           & testA             & testA           \\
                           &  param.$\downarrow$  &  param.$\downarrow$      & param.$\downarrow$ & (G)$\downarrow$  & FPS$\uparrow$    &  FPS$\uparrow$  & time$\downarrow$  & Acc.$\uparrow$  \\
    \midrule
    TransVG \cite{deng2021transvg}          & 150M &  21M  & 171M  & 214  & 22.8 &  59.6  &   95s & 82.7 \\
    QRNet \cite{qrnet}                      & 252M &  21M  & 273M  & 540  &  9.4 &  50.9  &  111s & 85.9 \\ 
    TransVG++$^\dag$ \cite{transvg++}       & 161M &  10M  & 171M  & 396  &  2.6 &   8.7  &  644s & 88.4 \\  
    MDETR \cite{kamath2021mdetr}            & 150M & 135M  & 185M  & 642  &  4.7 &  19.9  &  283s & 89.6 \\  
    Grounding-DINO \cite{liu2023grounding}  & 156M &  15M  & 172M  & 464  &  --  &   8.3  &  681s & 91.8 \\  
    \midrule
    \rowcolor{cyan!03}
    \textbf{UniRef (Ours)}     & \textbf{147M} & \textbf{1.7M}  & \textbf{149M}  & \textbf{162}  & \textbf{55.8} &  \textbf{83.2}  &   \textbf{68s} & \textbf{94.3} \\  
    \bottomrule
\end{tabular}%
}
\end{center}
\label{tab:cost}
\vspace{-10pt}	
\end{table*}

\subsection{Computational costs analysis compared with SoTA methods}
\label{sup_subsec:computational_costs_analysis}

In this paper, we highlight two significant advantages of our model architecture over other frameworks: (a) Instead of using a Transformer to fuse visual and language features, we only employ a simple lightweight task head; (b) Our one-tower architecture eliminates the need for early interaction techniques in the backbone network, thereby reducing the computational complexity of the model.

We compare the energy efficiency of our model with several well-known SoTA works on the REC task from various perspectives, including the number of parameters, computational complexity (FLOPs), inference speed (FPS), and test time (s). As can be seen from \cref{tab:cost}, due to the simplification of our model's structure, the number of parameters and the calculation complexity are significantly lower than other well-known models. Specifically, our feature fusion and grounding head module only require 1.7M parameters, while other methods use 20M, meaning we only have about 8.5$\%$ of their parameter count. Additionally, our computation is only 34.9$\%$ of Grounding-DINO and 25.2$\%$ of MDETR. Moreover, our inference speed is 10 $\times$ faster than Grounding-DINO and TransVG++ (the speed also related to the image size used by the model). Despite these advantages, thanks to the modality-shared feature space, we outperform all these well-known works.

\begin{table*}[t]
    \footnotesize
    \caption{Complete ablation study of MRefM using our OneRef-base model in REC task on both single-dataset fine-tuning setting and mixup intermediate pre-training setting.(Acc@0.5($\%$))}
    \centering
    \resizebox{0.9\columnwidth}{!}{%
    \begin{tabular}{c|c|c|ccc|ccc|cc}
    \toprule
    \multirow{2}[1]{*}{MIM} & \multirow{2}[1]{*}{MLM} & image masking & \multicolumn{3}{c|}{RefCOCO} & \multicolumn{3}{c|}{RefCOCO+} & \multicolumn{2}{c}{RefCOCOg}  \\
        &     & strategy   &  val  & testA & testB     & val  & testA & testB        &   val  & test                 \\
    \midrule
    \multicolumn{11}{l}{\textbf{Single-dataset fine-tuning setting:}}       \\ 
    
    \XSolidBrush    &   \XSolidBrush  & \XSolidBrush       & 85.23  & 88.13 & 83.82 & 78.56 & 83.36 & 71.72 & 80.41 & 80.52  \\
    vanilla         &   vanilla       &  block-wise        & 85.75 & 88.86 & 83.43 & 78.47 & 84.27 & 71.66 & 81.06 & 81.19  \\
    \rowcolor{cyan!03}
    Ref MIM         &   Ref MLM       &  referring-aware   & \textbf{88.75} & \textbf{90.95} & \textbf{85.34} & \textbf{80.43}  & \textbf{86.46} & \textbf{74.26} & \textbf{83.68} & \textbf{83.52}  \\
    \midrule
    \multicolumn{11}{l}{\textbf{Dataset-mixed intermediate pre-training setting: (main) }}       \\ 
    \XSolidBrush    &   \XSolidBrush  & \XSolidBrush     & 85.23 & 88.13 & 83.82 & 78.56 & 83.36 & 71.72 & 80.41 & 80.52  \\
    vanilla         &   vanilla       &  random          & 86.60 & 89.86 & 84.96 & 79.68 & 84.59 & 72.11 & 81.35 & 81.11  \\
    vanilla         &   vanilla       &  referring-aware & 86.71 & 90.58 & 85.33 & 80.06 & 85.77 & 73.96 & 81.96 & 82.16  \\
    Ref MIM         &   vanilla       &  referring-aware & 88.86 & 92.12 & 86.89 & 83.64 & 88.26 & 76.58 & 83.55 & 85.86  \\
    vanilla         &   Ref MLM       &  referring-aware & 87.26 & 91.68 & 86.37 & 81.52 & 86.87 & 75.93 & 82.88 & 84.32  \\
    Ref MIM         &   Ref MLM       &  random          & 90.56 & 93.55 & 88.23 & 85.08 & 89.12 & 78.56 & 85.57 & 86.89  \\    
    Ref MIM         &   Ref MLM       &  block-wise      & 90.07 & 93.32 & 88.21 & 84.55 & 88.83 & 77.98 & 84.71 & 86.69  \\  
    \rowcolor{cyan!03}
    Ref MIM         &   Ref MLM       &  referring-aware & \textbf{91.89} & \textbf{94.31} & \textbf{88.58} & \textbf{86.38}  & \textbf{90.38} & \textbf{79.47} & \textbf{86.82} & \textbf{87.32}  \\
    \bottomrule
    \end{tabular}%
    }
    \label{tab:ablation_mrm_full}%
    \vspace{-7pt}  
\end{table*}%

\begin{table*}[t!]
    \footnotesize
    \caption{Ablation study of the mask ratio in referring-aware dynamic masking strategy on RefCOCOg(val) dataset.}
    \vspace{-3pt}
    \begin{center}
    \resizebox{0.55\columnwidth}{!}{%
    \begin{tabular}{c|c|c||c|c|c}
        \toprule
        \multicolumn{2}{c}{mask ratio}   &  RefCOCOg    &   \multicolumn{2}{c}{mask ratio}   &  RefCOCOg     \\
        $\beta$     &  $\gamma$   &   Acc@0.5($\%$)   &  $\beta$     &  $\gamma$  &  Acc@0.5($\%$) \\
        \midrule
        0.20  & 0.75 & 83.40 &  0.35 &  0.50        &  85.98    \\
        0.30  & 0.75 & 85.01 &  0.35 &  0.60        &  86.32    \\
        \textbf{0.35}  & \textbf{0.75} & \textbf{86.82} & 0.35 &  0.70        &  86.62    \\
        0.40  & 0.75 & 86.62 &  \textbf{0.35} &  \textbf{0.75}        &  \textbf{86.82}    \\
        0.50  & 0.75 & 86.23 &  0.35 &  0.80        &  86.19    \\
        0.60  & 0.75 & 84.72 &  0.35 &  0.85        &  85.56    \\
        0.70  & 0.75 & 83.39 &  0.35 &  0.90        &  84.87    \\
        \bottomrule
    \end{tabular}%
    }
    \end{center}
    \label{tab:dynamic_mask}
\end{table*}%

\begin{figure}[t!]
 \centering
   \includegraphics[width=0.97\linewidth]{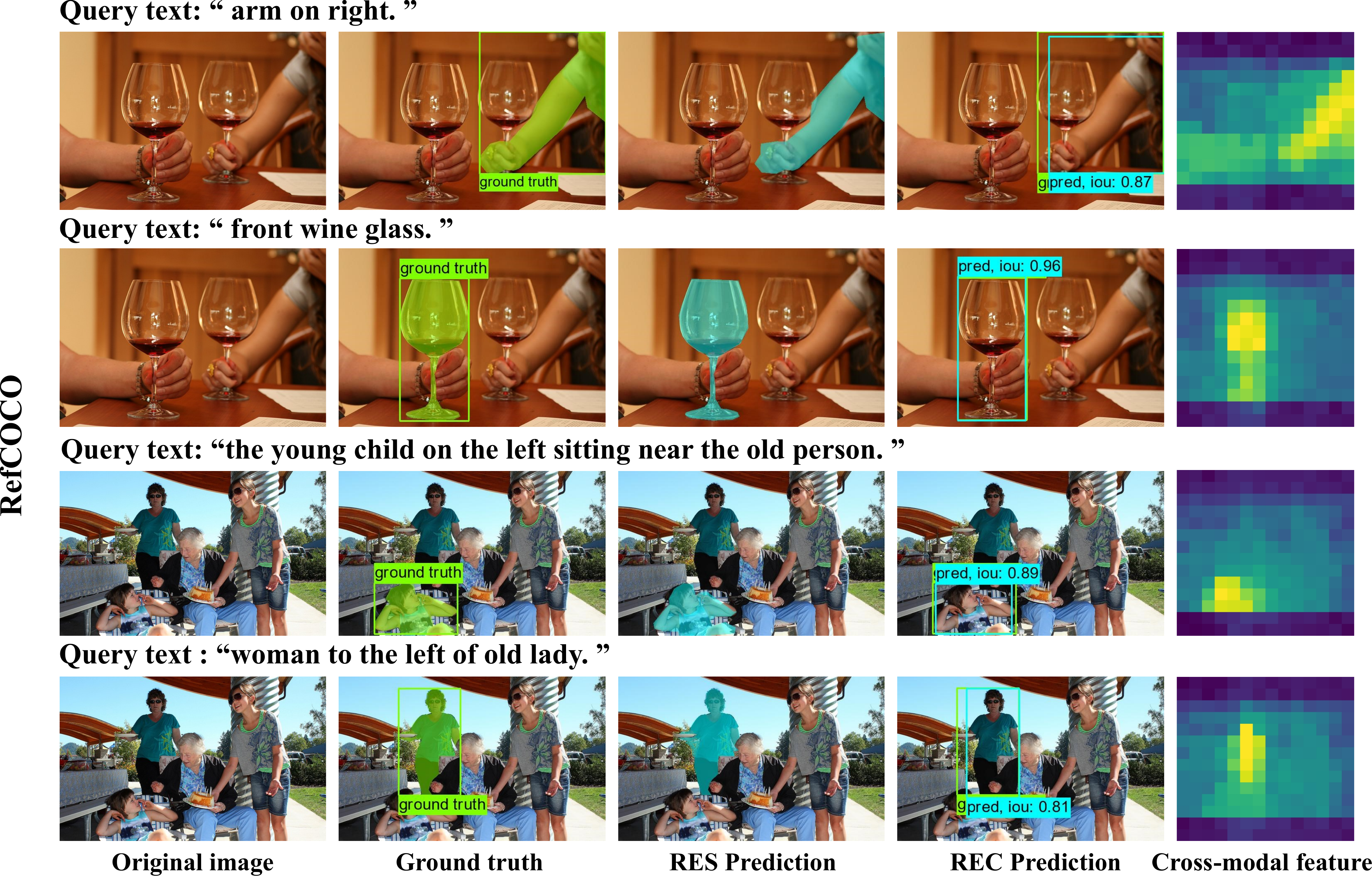}
   \vspace{-5pt}
   \caption{Qualitative results of our OneRef framework on the RefCOCO-val split. Each example shows two different query texts. From left to right: the original input image, the ground truth with box and segmentation mask (in \textcolor{green}{green}), the RES prediction of OneRef (in \textcolor{cyan}{cyan}), the REC prediction of OneRef (in \textcolor{cyan}{cyan}), and the cross-modal feature.}
    \vspace{-7pt}
   \label{fig:visual_unc}
\end{figure}

\begin{figure}[!h]
 \centering
   \includegraphics[width=0.97\linewidth]{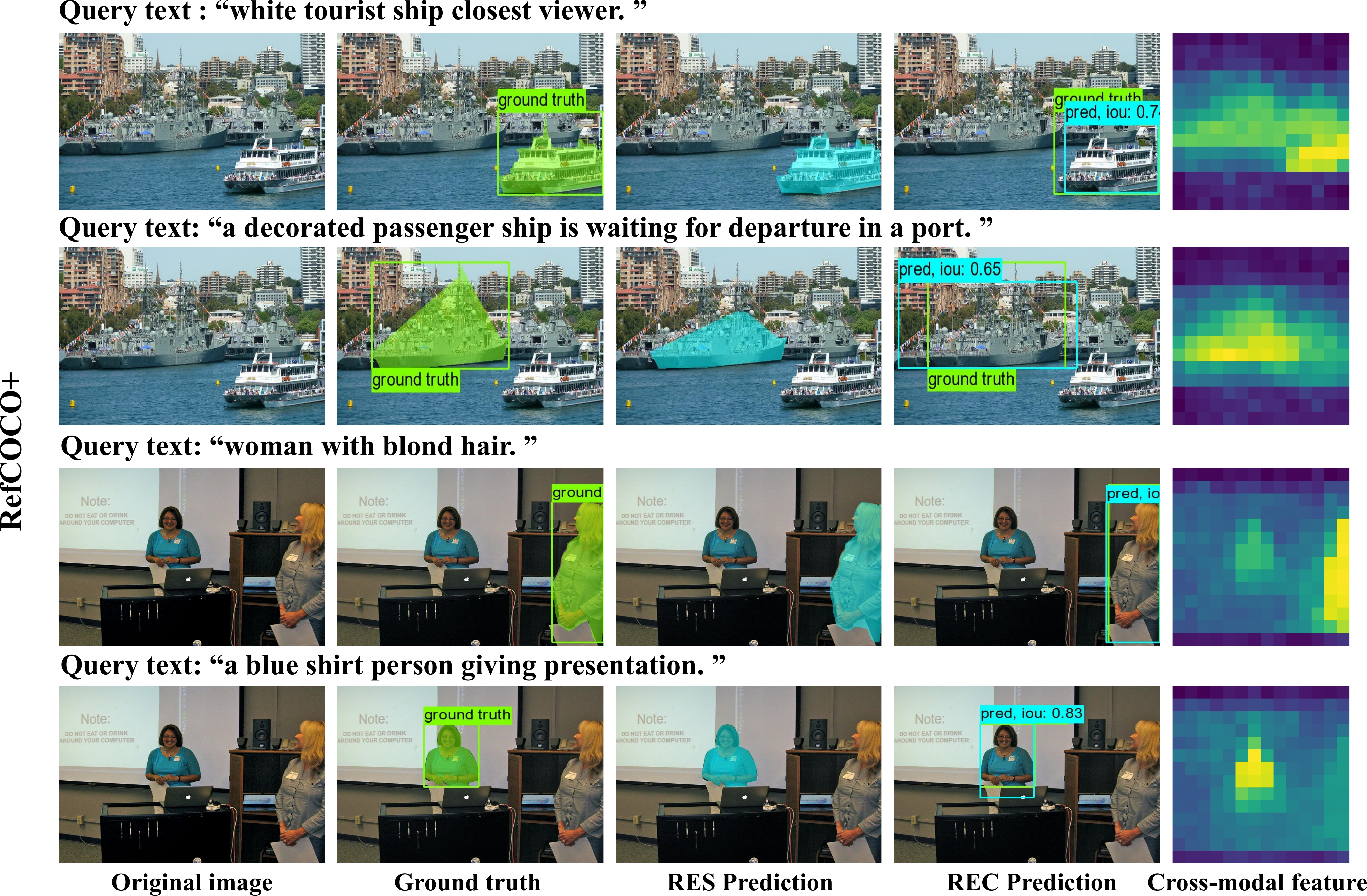}
   \vspace{-4pt}
   \caption{Qualitative results on the RefCOCO+-val dataset. The annotation is the same as \cref{fig:visual_unc}.}
   \vspace{9pt}
   \label{fig:visual_unc+}
\end{figure}

\begin{figure}[!h]
 \centering
   \includegraphics[width=0.97\linewidth]{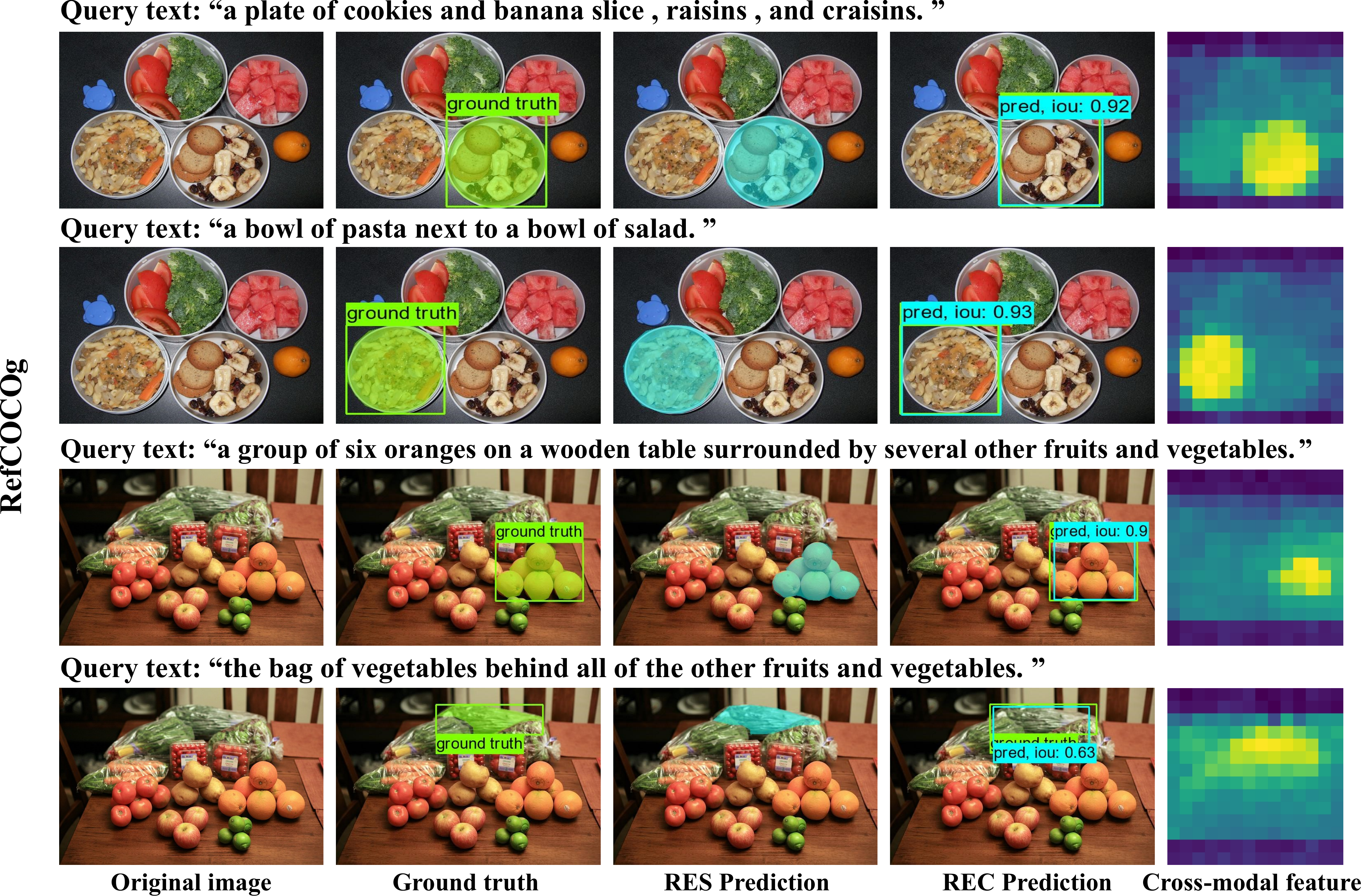}
   \vspace{-4pt}
   \caption{Qualitative results on the RefCOCOg-val dataset. The annotation is the same as \cref{fig:visual_unc}.}
    \vspace{7pt}
   \label{fig:visual_gref_umd}
\end{figure}

\subsection{Complete ablation study of MRefM on single-dataset fine-tuning and mixup pretrain settings}
\label{sup_subsec:Complete_ablation_study_of_MRefM}

The complete ablation results of MRefM on both single-dataset fine-tuning and mixup pretrain settings are provided in \cref{tab:ablation_mrm_full}, which serves as a supplement to \cref{tab:ablation_mrm} in the main text. In the table, the masking ratio is set to 0.4 when using block-wise or random masking strategies.

\subsection{Ablation study of the mask ratio in referring-aware dynamic masking strategy}
\label{sup_subsec:Complete_ablation_study_of_dynamic_masking}

As depicted in \cref{tab:dynamic_mask}, we conducted ablation experiments on the mask ratio within our proposed referring-aware dynamic image masking strategy. It is observed that while a high mask ratio of 0.75 is employed for the pixel reconstruction of MAE \cite{he2022masked}, achieving better results for BEiT's feature reconstruction requires a mask ratio ranging from approximately 0.4 to 0.45. In our proposed approach, favorable outcomes can be attained by setting $\beta$ and $\gamma$ to 0.35 and 0.75, respectively; where $\beta$ represents the mask ratio beyond the referred region and $\gamma$ denotes the mask ratio within it. Experimental statistics show that our entire mask rate $\alpha$ in each sample is about $0.4\sim0.5$.

\section{Visualization of the results}
\label{sup_sec:visualize}

As shown in \cref{fig:visual_unc}, \cref{fig:visual_unc+}, and \cref{fig:visual_gref_umd}, we present the qualitative grounding and referring segmentation results with several relatively challenging examples. Each example shows two different query texts. The cross-modal features are obtained by the cosine similarity between the \texttt{[SEP]} language token and the vision tokens on REC transfer model of OneRef-B. These results demonstrate the strong semantic comprehension capability of our OneRef model in complex text understanding and cross-modal grounding.

\section{Further discussions}
\label{sup_sec:further_discuss}

\subsection{Limitations}
\label{sup_subsec:Limitations}

Firstly, despite achieving remarkable grounding and segmentation results, the pre-training in this paper solely relies on the comparatively limited RefC dataset, as opposed to other studies with larger datasets.

Secondly, the MRefM paradigm necessitates additional referential bounding boxes as supervised data compared to self-supervised pre-training. Therefore, we explore the potential of unsupervised pre-training for MRefM. However, when utilizing image-text pairs obtained from web crawling, there is no guarantee that the referred regions will exhibit strong correlation with the text due to many texts describing the entire image. This aspect introduces certain challenges and biases during large-scale pre-training of MRefM. Consequently, this paper should serve as an inspiration for subsequent researchers to propose more convenient plug-and-play modeling methods.

\subsection{Broader impacts}
\label{sup_subsec:Broader_impacts}

OneRef demonstrates strong grounding and referring segmentation capabilities, while MRefM represents a novel modeling paradigm for referential relationship. This facilitates users to easily utilize our model (\eg, OneRef-L) for their own needs by simply providing some appropriate text queries. However, this also raises concerns about how our OneRef models with a strong understanding capabilities could be used inappropriately in the community, such as for large-scale illegal video surveillance. The open-set grounding capabilities could be manipulated through specialized textual cues to facilitate targeted detection or human tracking instead of generic ones. This manipulation could introduce biases in the detector and result in unfair predictions.


\newpage

\section*{NeurIPS Paper Checklist}

\begin{enumerate}

\item {\bf Claims}
    \item[] Question: Do the main claims made in the abstract and introduction accurately reflect the paper's contributions and scope?
    \item[] Answer: \answerYes{} 
    \item[] Justification: The abstract and introduction clearly state the claims made, including the contributions made in the paper and important assumptions and limitations.
    \item[] Guidelines:
    \begin{itemize}
        \item The answer NA means that the abstract and introduction do not include the claims made in the paper.
        \item The abstract and/or introduction should clearly state the claims made, including the contributions made in the paper and important assumptions and limitations. A No or NA answer to this question will not be perceived well by the reviewers. 
        \item The claims made should match theoretical and experimental results, and reflect how much the results can be expected to generalize to other settings. 
        \item It is fine to include aspirational goals as motivation as long as it is clear that these goals are not attained by the paper. 
    \end{itemize}

\item {\bf Limitations}
    \item[] Question: Does the paper discuss the limitations of the work performed by the authors?
    \item[] Answer: \answerYes{} 
    \item[] Justification: We have discussed the limitations of this work in the \cref{sup_subsec:Limitations}.
    \item[] Guidelines:
    \begin{itemize}
        \item The answer NA means that the paper has no limitation while the answer No means that the paper has limitations, but those are not discussed in the paper. 
        \item The authors are encouraged to create a separate "Limitations" section in their paper.
        \item The paper should point out any strong assumptions and how robust the results are to violations of these assumptions (e.g., independence assumptions, noiseless settings, model well-specification, asymptotic approximations only holding locally). The authors should reflect on how these assumptions might be violated in practice and what the implications would be.
        \item The authors should reflect on the scope of the claims made, e.g., if the approach was only tested on a few datasets or with a few runs. In general, empirical results often depend on implicit assumptions, which should be articulated.
        \item The authors should reflect on the factors that influence the performance of the approach. For example, a facial recognition algorithm may perform poorly when image resolution is low or images are taken in low lighting. Or a speech-to-text system might not be used reliably to provide closed captions for online lectures because it fails to handle technical jargon.
        \item The authors should discuss the computational efficiency of the proposed algorithms and how they scale with dataset size.
        \item If applicable, the authors should discuss possible limitations of their approach to address problems of privacy and fairness.
        \item While the authors might fear that complete honesty about limitations might be used by reviewers as grounds for rejection, a worse outcome might be that reviewers discover limitations that aren't acknowledged in the paper. The authors should use their best judgment and recognize that individual actions in favor of transparency play an important role in developing norms that preserve the integrity of the community. Reviewers will be specifically instructed to not penalize honesty concerning limitations.
    \end{itemize}

\item {\bf Theory Assumptions and Proofs}
    \item[] Question: For each theoretical result, does the paper provide the full set of assumptions and a complete (and correct) proof?
    \item[] Answer: \answerYes{} 
    \item[] Justification: We have provided the full set of assumptions and a complete proof in \cref{3.0-method} of the main text.
    \item[] Guidelines:
    \begin{itemize}
        \item The answer NA means that the paper does not include theoretical results. 
        \item All the theorems, formulas, and proofs in the paper should be numbered and cross-referenced.
        \item All assumptions should be clearly stated or referenced in the statement of any theorems.
        \item The proofs can either appear in the main paper or the supplemental material, but if they appear in the supplemental material, the authors are encouraged to provide a short proof sketch to provide intuition. 
        \item Inversely, any informal proof provided in the core of the paper should be complemented by formal proofs provided in appendix or supplemental material.
        \item Theorems and Lemmas that the proof relies upon should be properly referenced. 
    \end{itemize}

    \item {\bf Experimental Result Reproducibility}
    \item[] Question: Does the paper fully disclose all the information needed to reproduce the main experimental results of the paper to the extent that it affects the main claims and/or conclusions of the paper (regardless of whether the code and data are provided or not)?
    \item[] Answer: \answerYes{} 
    \item[] Justification: We have fully disclosed all the information needed to reproduce the main experimental results of the paper in \cref{4.0-experiments} of the main text and \cref{sup_sec:imp_details}.
    \item[] Guidelines:
    \begin{itemize}
        \item The answer NA means that the paper does not include experiments.
        \item If the paper includes experiments, a No answer to this question will not be perceived well by the reviewers: Making the paper reproducible is important, regardless of whether the code and data are provided or not.
        \item If the contribution is a dataset and/or model, the authors should describe the steps taken to make their results reproducible or verifiable. 
        \item Depending on the contribution, reproducibility can be accomplished in various ways. For example, if the contribution is a novel architecture, describing the architecture fully might suffice, or if the contribution is a specific model and empirical evaluation, it may be necessary to either make it possible for others to replicate the model with the same dataset, or provide access to the model. In general. releasing code and data is often one good way to accomplish this, but reproducibility can also be provided via detailed instructions for how to replicate the results, access to a hosted model (e.g., in the case of a large language model), releasing of a model checkpoint, or other means that are appropriate to the research performed.
        \item While NeurIPS does not require releasing code, the conference does require all submissions to provide some reasonable avenue for reproducibility, which may depend on the nature of the contribution. For example
        \begin{enumerate}
            \item If the contribution is primarily a new algorithm, the paper should make it clear how to reproduce that algorithm.
            \item If the contribution is primarily a new model architecture, the paper should describe the architecture clearly and fully.
            \item If the contribution is a new model (e.g., a large language model), then there should either be a way to access this model for reproducing the results or a way to reproduce the model (e.g., with an open-source dataset or instructions for how to construct the dataset).
            \item We recognize that reproducibility may be tricky in some cases, in which case authors are welcome to describe the particular way they provide for reproducibility. In the case of closed-source models, it may be that access to the model is limited in some way (e.g., to registered users), but it should be possible for other researchers to have some path to reproducing or verifying the results.
        \end{enumerate}
    \end{itemize}

\item {\bf Open access to data and code}
    \item[] Question: Does the paper provide open access to the data and code, with sufficient instructions to faithfully reproduce the main experimental results, as described in supplemental material?
    \item[] Answer: \answerNo{} 
    \item[] Justification: We answer No mainly for the following acceptable reasons: (1) The data we use are all publicly available and have been detailedly introduced in the paper. Researchers can acquire the data according to the provided reference information. (2) Due to time constraints, we were unable to compile and submit the anonymous code at the time of submission. However, as stated in the abstract, all models and code will be promptly made public after the decision is reached on this paper. (3) The implementation details of our work have been thoroughly explained in both the main text and supplementary materials. Even without publicly available code, researchers can reproduce it based on the information given in this paper.
    \item[] Guidelines:
    \begin{itemize}
        \item The answer NA means that paper does not include experiments requiring code.
        \item Please see the NeurIPS code and data submission guidelines (\url{https://nips.cc/public/guides/CodeSubmissionPolicy}) for more details.
        \item While we encourage the release of code and data, we understand that this might not be possible, so “No” is an acceptable answer. Papers cannot be rejected simply for not including code, unless this is central to the contribution (e.g., for a new open-source benchmark).
        \item The instructions should contain the exact command and environment needed to run to reproduce the results. See the NeurIPS code and data submission guidelines (\url{https://nips.cc/public/guides/CodeSubmissionPolicy}) for more details.
        \item The authors should provide instructions on data access and preparation, including how to access the raw data, preprocessed data, intermediate data, and generated data, etc.
        \item The authors should provide scripts to reproduce all experimental results for the new proposed method and baselines. If only a subset of experiments are reproducible, they should state which ones are omitted from the script and why.
        \item At submission time, to preserve anonymity, the authors should release anonymized versions (if applicable).
        \item Providing as much information as possible in supplemental material (appended to the paper) is recommended, but including URLs to data and code is permitted.
    \end{itemize}

\item {\bf Experimental Setting/Details}
    \item[] Question: Does the paper specify all the training and test details (e.g., data splits, hyperparameters, how they were chosen, type of optimizer, etc.) necessary to understand the results?
    \item[] Answer: \answerYes{} 
    \item[] Justification: We have specified all the training and test details in the \cref{4.0-experiments} of the main text and \cref{sup_subsec:intro_of_dataset}.
    \item[] Guidelines:
    \begin{itemize}
        \item The answer NA means that the paper does not include experiments.
        \item The experimental setting should be presented in the core of the paper to a level of detail that is necessary to appreciate the results and make sense of them.
        \item The full details can be provided either with the code, in appendix, or as supplemental material.
    \end{itemize}

\item {\bf Experiment Statistical Significance}
    \item[] Question: Does the paper report error bars suitably and correctly defined or other appropriate information about the statistical significance of the experiments?
    \item[] Answer: \answerNo{} 
    \item[] Justification: The error bars are not reported because it would be too computationally expensive. Besides, it is not crucial for interpreting the experimental results in this task topic.
    \item[] Guidelines:
    \begin{itemize}
        \item The answer NA means that the paper does not include experiments.
        \item The authors should answer "Yes" if the results are accompanied by error bars, confidence intervals, or statistical significance tests, at least for the experiments that support the main claims of the paper.
        \item The factors of variability that the error bars are capturing should be clearly stated (for example, train/test split, initialization, random drawing of some parameter, or overall run with given experimental conditions).
        \item The method for calculating the error bars should be explained (closed form formula, call to a library function, bootstrap, etc.)
        \item The assumptions made should be given (e.g., Normally distributed errors).
        \item It should be clear whether the error bar is the standard deviation or the standard error of the mean.
        \item It is OK to report 1-sigma error bars, but one should state it. The authors should preferably report a 2-sigma error bar than state that they have a 96\% CI, if the hypothesis of Normality of errors is not verified.
        \item For asymmetric distributions, the authors should be careful not to show in tables or figures symmetric error bars that would yield results that are out of range (e.g. negative error rates).
        \item If error bars are reported in tables or plots, The authors should explain in the text how they were calculated and reference the corresponding figures or tables in the text.
    \end{itemize}

\item {\bf Experiments Compute Resources}
    \item[] Question: For each experiment, does the paper provide sufficient information on the computer resources (type of compute workers, memory, time of execution) needed to reproduce the experiments?
    \item[] Answer: \answerYes{} 
    \item[] Justification: We provide sufficient information on the computer resources in \cref{sup_sec:imp_details}.
    \item[] Guidelines:
    \begin{itemize}
        \item The answer NA means that the paper does not include experiments.
        \item The paper should indicate the type of compute workers CPU or GPU, internal cluster, or cloud provider, including relevant memory and storage.
        \item The paper should provide the amount of compute required for each of the individual experimental runs as well as estimate the total compute. 
        \item The paper should disclose whether the full research project required more compute than the experiments reported in the paper (e.g., preliminary or failed experiments that didn't make it into the paper). 
    \end{itemize}
    
\item {\bf Code Of Ethics}
    \item[] Question: Does the research conducted in the paper conform, in every respect, with the NeurIPS Code of Ethics \url{https://neurips.cc/public/EthicsGuidelines}?
    \item[] Answer: \answerYes{} 
    \item[] Justification: We have conformed in every respect with the NeurIPS Code of Ethics.
    \item[] Guidelines:
    \begin{itemize}
        \item The answer NA means that the authors have not reviewed the NeurIPS Code of Ethics.
        \item If the authors answer No, they should explain the special circumstances that require a deviation from the Code of Ethics.
        \item The authors should make sure to preserve anonymity (e.g., if there is a special consideration due to laws or regulations in their jurisdiction).
    \end{itemize}

\item {\bf Broader Impacts}
    \item[] Question: Does the paper discuss both potential positive societal impacts and negative societal impacts of the work performed?
    \item[] Answer: \answerYes{} 
    \item[] Justification: We have discussed both potential positive societal impacts and negative societal impacts of the work performed in \cref{sup_subsec:Broader_impacts}.
    \item[] Guidelines:
    \begin{itemize}
        \item The answer NA means that there is no societal impact of the work performed.
        \item If the authors answer NA or No, they should explain why their work has no societal impact or why the paper does not address societal impact.
        \item Examples of negative societal impacts include potential malicious or unintended uses (e.g., disinformation, generating fake profiles, surveillance), fairness considerations (e.g., deployment of technologies that could make decisions that unfairly impact specific groups), privacy considerations, and security considerations.
        \item The conference expects that many papers will be foundational research and not tied to particular applications, let alone deployments. However, if there is a direct path to any negative applications, the authors should point it out. For example, it is legitimate to point out that an improvement in the quality of generative models could be used to generate deepfakes for disinformation. On the other hand, it is not needed to point out that a generic algorithm for optimizing neural networks could enable people to train models that generate Deepfakes faster.
        \item The authors should consider possible harms that could arise when the technology is being used as intended and functioning correctly, harms that could arise when the technology is being used as intended but gives incorrect results, and harms following from (intentional or unintentional) misuse of the technology.
        \item If there are negative societal impacts, the authors could also discuss possible mitigation strategies (e.g., gated release of models, providing defenses in addition to attacks, mechanisms for monitoring misuse, mechanisms to monitor how a system learns from feedback over time, improving the efficiency and accessibility of ML).
    \end{itemize}
    
\item {\bf Safeguards}
    \item[] Question: Does the paper describe safeguards that have been put in place for responsible release of data or models that have a high risk for misuse (e.g., pretrained language models, image generators, or scraped datasets)?
    \item[] Answer: \answerNA{} 
    \item[] Justification: Our paper poses no such risks.
    \item[] Guidelines:
    \begin{itemize}
        \item The answer NA means that the paper poses no such risks.
        \item Released models that have a high risk for misuse or dual-use should be released with necessary safeguards to allow for controlled use of the model, for example by requiring that users adhere to usage guidelines or restrictions to access the model or implementing safety filters. 
        \item Datasets that have been scraped from the Internet could pose safety risks. The authors should describe how they avoided releasing unsafe images.
        \item We recognize that providing effective safeguards is challenging, and many papers do not require this, but we encourage authors to take this into account and make a best faith effort.
    \end{itemize}

\item {\bf Licenses for existing assets}
    \item[] Question: Are the creators or original owners of assets (e.g., code, data, models), used in the paper, properly credited and are the license and terms of use explicitly mentioned and properly respected?
    \item[] Answer: \answerYes{} 
    \item[] Justification: We have cited the original paper (\eg, BEiT-3 \cite{beit3}) that produced the code package or dataset.
    \item[] Guidelines:
    \begin{itemize}
        \item The answer NA means that the paper does not use existing assets.
        \item The authors should cite the original paper that produced the code package or dataset.
        \item The authors should state which version of the asset is used and, if possible, include a URL.
        \item The name of the license (e.g., CC-BY 4.0) should be included for each asset.
        \item For scraped data from a particular source (e.g., website), the copyright and terms of service of that source should be provided.
        \item If assets are released, the license, copyright information, and terms of use in the package should be provided. For popular datasets, \url{paperswithcode.com/datasets} has curated licenses for some datasets. Their licensing guide can help determine the license of a dataset.
        \item For existing datasets that are re-packaged, both the original license and the license of the derived asset (if it has changed) should be provided.
        \item If this information is not available online, the authors are encouraged to reach out to the asset's creators.
    \end{itemize}

\item {\bf New Assets}
    \item[] Question: Are new assets introduced in the paper well documented and is the documentation provided alongside the assets?
    \item[] Answer: \answerNA{} 
    \item[] Justification: Our paper does not release new assets.
    \item[] Guidelines:
    \begin{itemize}
        \item The answer NA means that the paper does not release new assets.
        \item Researchers should communicate the details of the dataset/code/model as part of their submissions via structured templates. This includes details about training, license, limitations, etc. 
        \item The paper should discuss whether and how consent was obtained from people whose asset is used.
        \item At submission time, remember to anonymize your assets (if applicable). You can either create an anonymized URL or include an anonymized zip file.
    \end{itemize}

\item {\bf Crowdsourcing and Research with Human Subjects}
    \item[] Question: For crowdsourcing experiments and research with human subjects, does the paper include the full text of instructions given to participants and screenshots, if applicable, as well as details about compensation (if any)? 
    \item[] Answer: \answerNA{} 
    \item[] Justification: This paper does not involve crowdsourcing nor research with human subjects.
    \item[] Guidelines:
    \begin{itemize}
        \item The answer NA means that the paper does not involve crowdsourcing nor research with human subjects.
        \item Including this information in the supplemental material is fine, but if the main contribution of the paper involves human subjects, then as much detail as possible should be included in the main paper. 
        \item According to the NeurIPS Code of Ethics, workers involved in data collection, curation, or other labor should be paid at least the minimum wage in the country of the data collector. 
    \end{itemize}

\item {\bf Institutional Review Board (IRB) Approvals or Equivalent for Research with Human Subjects}
    \item[] Question: Does the paper describe potential risks incurred by study participants, whether such risks were disclosed to the subjects, and whether Institutional Review Board (IRB) approvals (or an equivalent approval/review based on the requirements of your country or institution) were obtained?
    \item[] Answer: \answerNA{} 
    \item[] Justification: Our paper does not involve crowdsourcing nor research with human subjects.
    \item[] Guidelines:
    \begin{itemize}
        \item The answer NA means that the paper does not involve crowdsourcing nor research with human subjects.
        \item Depending on the country in which research is conducted, IRB approval (or equivalent) may be required for any human subjects research. If you obtained IRB approval, you should clearly state this in the paper. 
        \item We recognize that the procedures for this may vary significantly between institutions and locations, and we expect authors to adhere to the NeurIPS Code of Ethics and the guidelines for their institution. 
        \item For initial submissions, do not include any information that would break anonymity (if applicable), such as the institution conducting the review.
    \end{itemize}

\end{enumerate}

\end{document}